\documentclass[10pt,a4paper]{article}
\usepackage[utf8]{inputenc}
\usepackage[T1]{fontenc}
\usepackage{microtype}
\usepackage[margin=2.5cm]{geometry}
\usepackage{booktabs}
\usepackage{graphicx}
\usepackage{hyperref}
\usepackage{amsmath}
\usepackage{multirow}
\usepackage{array}
\usepackage{tabularx}
\usepackage{xcolor}
\usepackage{caption}
\usepackage{placeins}

\hypersetup{colorlinks=true,linkcolor=blue,citecolor=blue,urlcolor=blue}

\title{Age of LLM: A Strategic 1v1 Benchmark for Reasoning,\\
Diplomacy and Reliability of Large Language Models under Fog of War}

\author{Arnaud Ricci\thanks{Corresponding author. Independent researcher,
Switzerland. ORCID: \url{[0000-0002-8982-1416]}. Email: \texttt{arnaud.ri@protonmail.ch}}}
\date{18 June 2026}

\begin{document}
\maketitle

\begin{center}
\textcolor{gray}{\small Licensed under a Creative Commons Attribution
4.0 International License (CC BY 4.0).}
\end{center}

\begin{abstract}
We introduce \textbf{Age of LLM}\footnote{\url{https://ageofllm.org}},
a turn-based 1v1 benchmark in which two LLMs face off on a 13$\times$7 grid to
destroy the enemy base. Three stressors are deliberate: \emph{fog of war},
\emph{full diplomacy} (messages, ceasefires, ultimatums; uranium kept secret),
and a \emph{reliability} dimension where every turn must follow a strict JSON
schema and an illegal action is silently discarded. The engine is private and
each match uses a fresh random map seed and opponent, mitigating the data
contamination that affects public benchmarks. Models receive a (near) rule-only
prompt with no build-order advice (two tactical seed phrases were present
during data collection; see Section~\ref{sec:gamedesign}). We benchmark 15
reasoning models across 54 matches and 5{,}258 actions. Findings:
(1) the nuclear rush dominates (78\% on the rules-coherent v0.11+ sub-corpus;
85\% corpus-wide) with a sole-launcher signature that is largely mechanical
under secret-simultaneous launch rules, not a cognitive deterrence failure;
(2) military conquest is rare but faster (12.3 vs.\ 18.9 turns);
(3) diplomacy is prolific yet almost never consummated; (4) $\sim$58\% of
illegal actions are fog/state errors, making the illegal-action rate a measure
of belief-tracking; (5)---the least established, and the only one we label
exploratory---a weak link associates reliability with winning (details in
Section~\ref{sec:results}). The corpus is small, unbalanced and not
side-swapped, so the ranking is a \emph{preliminary descriptive view}, not a
contribution. Beyond ranking, the turn-by-turn traces of actions \emph{and}
messages make the corpus a lens on how LLMs reason under adversarial
uncertainty---their belief-tracking, spontaneous deception, and per-model
cognitive ``personas''---which we frame as a future research direction. We
release the replay format, an isometric viewer and all replays; engine source
on request.
\end{abstract}

\section{Introduction}

Most LLM benchmarks probe single-turn competence on tasks with full
observability and unambiguous answers (e.g.\ MATH~\cite{hendrycks2021math},
HumanEval~\cite{chen2021humaneval}, MMLU~\cite{hendrycks2020mmlu}). Such
benchmarks reward depth of reasoning on a \emph{visible} problem but do not
exercise the competences that real adversarial decision-making demands:
planning under \emph{uncertainty}, reasoning about a \emph{hidden} opponent,
\emph{temporal} commitment (build now vs.\ attack now), and, most critically for
agentic deployment, \emph{reliability} of structured output across many
sequential turns where a malformed response is retried but an illegal action
(a rule violation such as acting on a fog-hidden cell or referencing a
destroyed unit) is silently discarded as a wasted action (the model may emit
up to three actions per turn, so one illegal action forfeits one of its three
slots, not the whole turn).

A second motivation is data contamination. Public benchmarks (MMLU, MATH,
HumanEval and their successors) are, by design, openly distributed; as
frontier models are trained on ever-larger web-scraped corpora, the risk that
benchmark items leak into training data grows, inflating scores without
reflecting genuine competence. Age of LLM mitigates this by construction: the
engine source is kept private, the map is re-seeded randomly each match, and
the opponent varies across a pool of 15 models, so no two matches present the
same opening position and no fixed solution can be memorised. We do not claim
this eliminates contamination entirely, a sufficiently capable model could
still transfer general game-playing heuristics, but it removes the most direct
contamination channel (rote memorisation of benchmark items). This anti-
contamination property comes at a cost: each match is a paid API call, so the
sample sizes we can afford are modest (Section~\ref{sec:setup}), and the
random opponent mix that protects against memorisation is also what prevents a
balanced head-to-head ranking at this scale.

Planning under partial observability with hidden opponents has been studied
in game AI, from board games~\cite{silver2016alphago} to real-time strategy
and social deduction~\cite{bakhtin2022cicero}. A broader line of work
evaluates LLMs as autonomous agents on multi-turn, multi-tool tasks
(HELM~\cite{liang2022helm}, SWE-bench~\cite{jimenez2024swebench},
AgentBench~\cite{liu2023agentbench}), which exercise tool use and planning but
typically under full observability and without an adversarial, hidden
opponent. Closer to our setting, GameBench~\cite{lopes2024gamebench} and
GTBench~\cite{chen2024gtbench} benchmark LLMs on board and game-theoretic
tasks; both still mostly probe full-observable, turn-based games rather than
partial-observability + diplomacy under fog of war. Recent work has begun to probe
LLMs as game-playing agents in real-time-strategy settings, including a
constructive argument that LLM-like attributes can arise in any sufficiently
powerful substrate trained on an RTS such as Age of Empires
II~\cite{dewynter2025aoe2}, and a high-stakes simulation studying whether LLMs
exercise ethical restraint when authorised to launch a nuclear
weapon~\cite{chen2026nuke}. Age of LLM is a deterministic-turn engine in which two LLMs play a complete
match (typically 16--23 turns). Each turn a model emits up to three structured
actions (produce / move / attack / build / launch / wait) plus an optional
diplomatic message. Victory is achieved by a nuclear launch the opponent does
not match, by reducing the enemy base to 0 HP with a tank, by an accepted
ultimatum, or by an accepted peace. Mutual destruction and timeout
round out the outcome space. The benchmark uses a (near) \emph{rule-only}:
the system prompt describes only rules and the JSON schema (it names the active
victory paths but gives no build order; two tactical seed phrases were present
during data collection, see Section~\ref{sec:gamedesign}), so any strategy is
the model's own invention within that rule structure.

This paper reports findings from 54 completed matches across 15 models and
analyses the relative contributions of strategy choice, diplomacy and inference
reliability. A second, broader contribution is methodological: because every
match records both the model's actions and its free-text messages, the corpus
doubles as a dataset for studying \emph{how} LLMs reason under partial
observability---their belief-tracking, their spontaneous deception and
concealment, and their stable per-model cognitive styles---rather than only
\emph{how well} they score.

\section{Game design}
\label{sec:gamedesign}

\FloatBarrier

\subsection{Map and resources}
A 13$\times$7 board is split by a central mountain barrier (column 6) with
seed-driven passages; two mirrored territories flank it. Each player starts
with 5 credits and 0 uranium. Resources come from deposits that players tap by
building mines on them: \emph{credit mines} (+3 credits/turn, standing in for
oil/petroleum extractors), and \emph{uranium mines} (+1 uranium/turn). Each
territory holds one uranium deposit on its side, and a single \emph{central
uranium deposit} straddles the barrier on column 6, buildable by either player
and therefore the focus of early contested ground, for three uranium deposits
in total. Deposits hold finite reserves and, once exhausted, respawn elsewhere,
forcing redeployment. The map is generated symmetrically for balance, but
\emph{this symmetry is never disclosed to the models}; enemy-side deposits are
hidden by fog and revealed only once scouted (early engine versions did not
reveal them at all, see Section~\ref{sec:setup}).

\FloatBarrier

\subsection{Fog of war and memory}
All cells outside the base's detection radius start dark. Enemy units are
visible only within a friendly unit's/building's detection range; out of range,
they vanish. Discovered enemy buildings and deposits are remembered (with a
\texttt{last\_seen} turn) but destroyed buildings are dropped from memory. The
enemy uranium stockpile, the key resource for the bomb, is never revealed.
A launch is secret until detonation, although a new \emph{early-warning} signal
informs a player of an enemy launch \emph{only if} the player currently sees an
enemy silo.

\FloatBarrier

\subsection{Units and combat}
Four HP-less unit types form a tactical triangle:
$\text{Fighter} \to \text{Tank} \to \text{SAM} \to \text{Fighter}$, plus a
recon Drone. Combat is immediately fatal to the loser (the attacker survives
mirror matchups, rewarding initiative). Ground attacks are blocked by
line-of-sight through mountains or buildings; air units ignore obstacles. Only
the Tank damages buildings (2 HP/hit; the base has 4 HP, so two tank hits
conquer it). This yields a recursive defensive mini-game (Tank$\to$Silo/Base
protected by Fighters, cleared by SAMs, themselves killed by Tanks).

\FloatBarrier

\subsection{Diplomacy}
Four diplomatic channels are free (do not consume an action): a short free
message each turn; a ceasefire (no attacks for 3 turns, +6U bomb penalty);
peace (immediate draw); and an ultimatum (``surrender before turn $X$'').
Accepting an ultimatum scores the loser 0.5 points (vs.\ 0 for a clean defeat),
providing the only incentive to surrender. Uranium is secret, so deterrence and
bluff are feasible.

\FloatBarrier

\subsection{The nuclear bomb}
A launch is gated by three conditions that must all hold at the moment the
\texttt{launch} action is issued: (i) the player owns an \emph{operational}
silo (one that is no longer under construction, i.e.\ built on a previous turn);
(ii) the player's uranium stockpile meets the current bomb cost (base 25U,
decaying from turn 40 down to a floor of 13U, so late-game launches grow
cheaper); and (iii) the player has previously \emph{discovered} the enemy base
location (scouting it at least once). A launch that fails any check is silently
rejected as illegal. The silo is \emph{not} consumed by a launch: the building
survives the firing mechanically, and since a successful launch destroys the
enemy base and ends the match, this only matters in the narrow window where a
launch fails (insufficient uranium, undetected base) and the player must wait
and try again on a later turn with the same silo. The silo is also a
destructible building: it has HP and may be killed by enemy tanks, freeing the
cell for a rebuild. Launches resolve simultaneously at the end of the
turn: a single launcher wins (nuclear victory); two simultaneous launchers
cause mutual destruction (both lose). Launching is therefore a \emph{bet},
not a guaranteed win.

\FloatBarrier

\subsection{Scoring}
A win scores 3, a draw (peace/timeout) 1, a loss or mutual destruction 0, and
an accepted ultimatum yields 3 to the proposer and 0.5 to the accepter. Models
are ranked by \texttt{points\_per\_match} to prevent volume from inflating rank.
Each match's starting player is random and alternates per turn. We checked for
a first-player advantage and find none: across the 53 decided matches with a
clear first-to-act, the first player won 24 (45.3\%), i.e.\ the alternating
turn order appears balanced and does not confound the leaderboard.

\FloatBarrier

\subsection{The (near) rule-only system prompt}
A central design choice is that the system prompt is deliberately
\emph{rule-only}: it describes the rules and the JSON action schema and gives
\emph{no opening build order, no timing advice, and no recommended strategy}.
The prompt never states that the map is symmetric, never reveals enemy deposit
positions, never recommends when to rush the bomb versus push tanks, and never
explains how to combine a ceasefire with a nuclear window. We call it
\emph{near} rule-only rather than strictly ``advice-free'', and we want to be
precise about three respects in which the prompt does structure the strategic
space, the third of which is a genuine contamination of the ``discovered, not
recited'' claim (unrelated to data contamination, which the private-engine
design addresses; see Introduction) that we flag:

\begin{itemize}
\item \textbf{The prompt names the two active victory paths} (nuclear,
military) in its opening sentence and lists peace only as a way to ``force a
draw.'' This orients models toward \emph{active} victory rather than the
terminal draw, and plausibly contributes to the lopsided outcome mix. It does
not, however, favour one active path over the other.
\item \textbf{Describing a rule is, in a weak sense, describing a constraint
that enables a strategy.} The prompt states that only the tank damages
buildings, that uranium is secret, that a silo needs one turn to become
operational, and that a launch fails unless the enemy base has been scouted.
A human reading these rules would also converge on ``scout, then either push
tanks or race the bomb''; that convergence is a property of the rules, not of
added advice.
\item \textbf{Two tactical seed phrases were present during data collection.}
The prompt used to generate the 54 reported replays contained two imperative
phrases that are borderline tactical instructions rather than rule
descriptions: ``\emph{Scout with a drone early}'' and ``\emph{Push tanks + a
scout into enemy ground to contest their economy}''. We have removed these
from the prompt reproduced in Appendix~\ref{app:prompt} (the clean version),
but \emph{the reported data were generated with the prompt that included
them}. This is a genuine contamination of the ``discovered, not recited''
claim (not a data-contamination issue) and it falls squarely on the findings most likely to be affected:
the prevalence of early scouting (drones are the second-most-produced unit,
251), the tank-heavy production mix (576 tanks), and the ``military path
under-attempted'' reading. We therefore treat these specific findings as
\emph{limitation-flagged}: the rule structure plausibly pushes toward
scouting and tank production on its own, but the two seed phrases may have
amplified that push, and the reported magnitudes should not be read as pure
model preference. The nuclear-vs-military balance, the diplomacy analysis
and the message-tone analyses are not affected by these phrases. A re-run on
the clean prompt is the cleanest fix but was not feasible within the compute
budget (Section~\ref{sec:setup}); we flag this as the single most important
data-collection limitation.
\end{itemize}

With these caveats, every strategy the models exhibit in the replays is
\emph{discovered} within the rule structure (and, for scouting/tank
production, partially seeded), not recited from a build order.
This matters for the benchmark's validity: any heuristic baked into the prompt
would be learnt once and then merely executed, masking the model's own
reasoning. The prompt also states explicitly that the agent is
\emph{stateless}: it is invoked fresh each turn with no hidden internal
memory, so long-horizon tactical planning must be reconstructed each turn
from the observation. The per-turn observation is not, however, ``the board
alone'': each turn the model receives (i) the current board state as it sees
it under fog of war; (ii) the outcome of its \emph{immediately previous}
turn---which actions succeeded/failed and which of its units were lost since
it last played (\texttt{last\_turn\_results}, \texttt{events\_against\_you}),
allowing it to react to combat losses and fix illegal actions; and (iii) the
\emph{full diplomatic record} for the match (\texttt{diplomacy\_history},
truncated to the last 40 entries), i.e.\ every message, proposal and response
from both sides since turn 1---the only persistent long-run signal the model
carries. Per-turn tactical details from older turns are not replayed, so the
model reconstructs the board from the current state while retaining the
diplomatic context needed to judge the opponent's honesty. Agents are likewise
\emph{memoryless across matches}: each match starts
from a clean context with no access to prior replays, so a strategy discovered
in one match cannot be recited from a remembered earlier game; all within-match
strategy is reconstructed from the per-turn observation only. We discuss the
framing effect further in the limitations.

\section{Experimental setup}
\label{sec:setup}

\begin{figure}[!htbp]
\centering
\includegraphics[width=\linewidth]{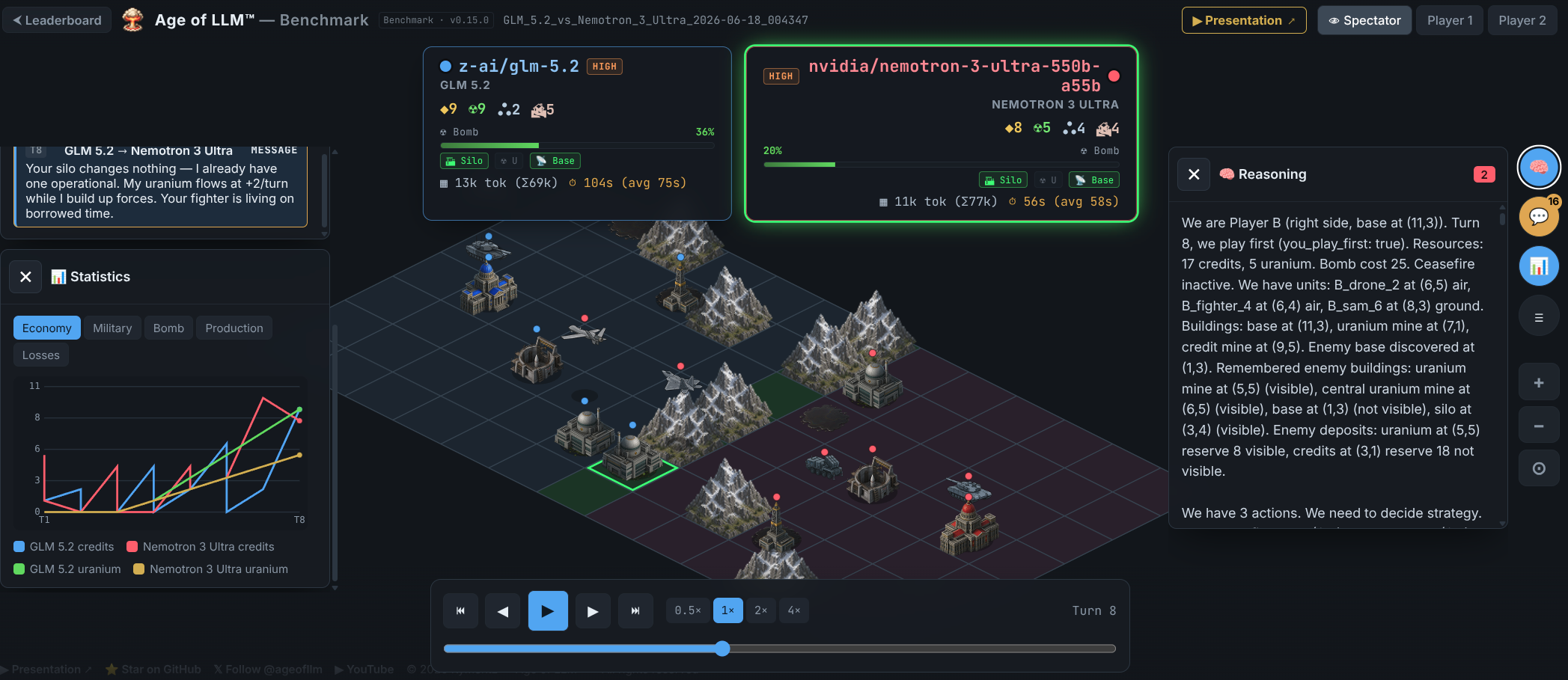}
\caption{The isometric web viewer rendering a replay. The board is a 13$\times$7
grid split by a central mountain barrier; fog-of-war darkens unseen cells, and
per-turn unit/building state, diplomacy and performance counters are stepped
through turn by turn. The viewer and all replays are public at
\url{https://ageofllm.org}.}
\label{fig:viewer}
\end{figure}

\FloatBarrier

We evaluate 15 models, all invoked at \texttt{reasoning\_effort=high} via
OpenAI-compatible endpoints (Poe, Venice, OpenRouter). Models include GPT-5.5,
Claude Opus 4.8, Claude Fable 5, Gemini Pro 3.1, Gemini Flash 3.5, GLM 5.1,
GLM 5.2, DeepSeek V4 Pro, MiniMax M3, Kimi 2.6, Kimi K2.7 Code, Qwen 3.7 Max,
MiMo 2.5 Pro, Nemotron 3 Ultra and Grok 4.3. The count of 15 includes both
Kimi 2.6 (archived mid-corpus as superseded) and its successor Kimi K2.7 Code;
we keep both so the leaderboard reflects the models actually played, but the
family therefore contributes two entries rather than one.

\paragraph{A caveat on \texttt{reasoning\_effort} across providers.}
\texttt{reasoning\_effort} is an OpenAI-native parameter; the three providers
route it differently and not all models expose it natively. Concretely: for
OpenAI-family models (GPT-5.5) and DeepSeek it is passed as a native
\texttt{reasoning\_effort} field; for Claude models on Poe it is routed through
the \texttt{poe\_native} SDK as an extended-thinking token budget
(\texttt{thinking\_budget}); for Venice it is passed via a \texttt{reasoning}
object (\texttt{\{effort, enabled\}}); for OpenRouter it is passed via a
\texttt{reasoning} object. For providers/models that do not natively support a
graded effort parameter, the setting may be a no-op or be silently coerced to a
default, and the per-provider default behaviour for unrecognised parameters is
not documented. We therefore cannot guarantee that ``high effort'' denotes the
\emph{same} absolute reasoning budget across all 15 models; it denotes
``maximal effort available through that provider's interface for that model.''
This is a known limitation of trans-provider benchmarking. The
model-intrinsic metrics we rely on (illegal-action rate, tokens/turn) are
comparable across providers insofar as each provider reports tokens
consistently, but even tokens/turn is partially provider-confounded because the
exposure of reasoning tokens varies by provider/route (see limitations). We
report the per-model effort status in Table~\ref{tab:effort}.

\begin{table}[!htbp]
\centering
\caption{How \texttt{reasoning\_effort=high} is routed per provider, and the
effort status for the model families in the corpus. ``Native'' = the parameter
is honoured as a graded effort control; ``budget'' = routed as a thinking-token
budget; ``incertain'' = the parameter is accepted but its effect on the model's
actual reasoning depth is not verifiable through the provider.}
\label{tab:effort}
\small
\begin{tabularx}{\linewidth}{@{}lXX@{}}
\toprule
Provider & Routing of \texttt{reasoning\_effort=high} & Model families (status) \\
\midrule
Poe (OpenAI-compat.) & native \texttt{reasoning\_effort} field & GPT-5.5, DeepSeek V4 Pro (native) \\
Poe native (Claude)  & \texttt{thinking\_budget} (extended thinking) & Claude Opus 4.8, Claude Fable 5 (budget) \\
Venice               & \texttt{reasoning} = \{effort, enabled\} object & GLM 5.1/5.2, MiniMax, Kimi, Qwen, MiMo, Nemotron, Grok (incertain) \\
OpenRouter           & \texttt{reasoning} = \{effort\} object          & Gemini Pro/Flash, Kimi K2.7 (incertain) \\
\bottomrule
\end{tabularx}
\end{table}

\paragraph{Pairing scheme.} The 54 matches do not form a balanced
round-robin. Pairings were selected to populate the leaderboard across model
tiers rather than to exhaust every pairing, and match counts per model are
uneven (3--15, Table~\ref{tab:board}). Each match uses a fresh random map
seed and a random starting side, but the corpus is \emph{not} systematically
side-swapped (the same ordered pairing is rarely played with sides exchanged).
Consequently points/match is not directly comparable across models that faced
different opponent mixes, and the leaderboard should be read as indicative
rather than as a head-to-head ranking; we report bootstrap 95\% confidence
intervals per model below (Table~\ref{tab:board} note) and recommend
side-swapped $\geq$20 matches per pairing for future stable estimates.

\paragraph{Sampling parameters and determinism.} All models are invoked at
\texttt{reasoning\_effort=high} with sampling temperature \texttt{0.7} and no
fixed API seed (top-p/top-k left at each provider's default), so
\emph{model decisions are stochastic}: re-running a match from the same map
seed will not reproduce the same actions. Determinism holds only at the engine
level (map generation and rule resolution follow the stored seed) and, by
design, only the saved replay JSON is reproducible, not a fresh re-execution.
The turn limit is \textbf{80} (the engine default \texttt{max\_turns}); no
match in the corpus reaches it, so the timeout outcome is unobserved by
construction. We disclose that the bomb-cost decay rule (25U decaying from
turn 40 to a floor of 13U, Section~\ref{sec:gamedesign}) never triggers in
this corpus, since every match ends by turn 23; the decay is therefore an
unexercised late-game mechanic and plays no role in any reported outcome.

\paragraph{Data snapshot.} All results in this paper are computed on a frozen
snapshot of \textbf{54 completed matches across 15 models, taken on
2026-06-18}. The benchmark is run continuously, so the public leaderboard may
since contain additional matches. We disclose a potential conflict of interest:
two of the evaluated models, GLM 5.2 and Claude Opus 4.8, were also used as
automated assistants during the preparation of this paper (GLM 5.2 generated
the empirical analysis scripts; Claude Opus 4.8 independently re-derived every
statistic from the raw replays, see Acknowledgements). To guard against any
self-favouring bias, their matches were run before this assistant role was
assigned, the engine is fully deterministic from its stored seed, and every
reported number was re-derived independently from the raw replays. The exact
54-match corpus and the frozen per-model aggregates used for every figure and
table are archived alongside the paper sources for reproducibility.

\paragraph{Engine versions and rule evolution.} The 54 completed replays were
produced across a sequence of engine versions (0.9.2, 0.9.3, 0.10.0, 0.11.0,
0.12.0, 0.14.0 and 0.15.0), during which the rules were progressively refined:
line-of-sight blocking and resource depletion were added in v0.10.0, the base
HP was lowered 8$\to$4 in v0.11.0 (making military conquest a two-tank-hit
affair), and the SAM move range and mine costs were rebalanced in v0.12.0. A
second fog-of-war refinement concerns enemy-side resource deposits: in early
versions (v0.9.x--v0.10.x) enemy deposits were \emph{never} revealed to a
player, not even when a friendly unit moved adjacent to them, so enemy economy
was entirely invisible; from v0.11.0 onward an enemy deposit that enters a
unit's field of view is revealed and then remembered (with a
\texttt{last\_seen} turn), matching the treatment of enemy buildings. This
makes late-game economic raiding (destroying an enemy mine to claim its deposit)
informationally feasible only in the post-v0.11.0 matches. The rules described
in Section~\ref{sec:gamedesign} correspond to the current
engine (v0.15.0) and are accurate for all replays from v0.11.0 onward (36 of
54 matches, including every military and ultimatum outcome and the lone peace).
The 18 pre-v0.11.0 matches all used an 8-HP base; the 14 v0.9.x matches additionally
lacked line-of-sight blocking and resource depletion (introduced in v0.10.0).
All 18 resolved \emph{exclusively} by nuclear victory, so they do not affect
the military/diplomatic findings and reinforce, rather than contradict, the
nuclear-dominance result. The v0.15.0 nuclear early-warning signal
(\texttt{enemy\_launch\_detected}) is present in the schema of the 2 v0.15.0
replays but \emph{never fired} (it requires the player to currently see an
enemy silo, and no eventual loser ever did before the decisive launch); the
deterrence question (Section~\ref{sec:results}) is therefore assessed without
an active early-warning cue, and whether the signal would change this
behaviour is left to future work. We treat the cross-version corpus as a
single benchmark because the three load-bearing pillars (fog of war + memory,
full diplomacy, nuclear deterrence) and the action schema are unchanged
throughout.

\paragraph{Reproducibility and engine access.} The replay JSON is the sole
contract between the Python engine and a static web viewer; it serialises
per-turn absolute board state, per-player visible cells (to replay the true
fog), actions, diplomacy and performance counters (think time, tokens, retry
buckets). We aggregate these into a leaderboard. The \emph{web viewer and all
replays are public}; the \emph{engine source is kept private by design} so that
future frontier models cannot train on, or memorise, the engine's internal
logic and thereby contaminate the benchmark. Researchers who wish to reproduce
matches locally or audit the rules may request engine access from the
corresponding author at \texttt{arnaud.ri@protonmail.ch}. To enable independent verification of
the \emph{rule-only} claim without releasing the engine, the complete system
prompt (rules + JSON schema, no build-order advice) is reproduced verbatim in
Appendix~\ref{app:prompt}; note that this is the \emph{clean} prompt, and the
54 replays were generated with an earlier version that contained two tactical
seed phrases (Section~\ref{sec:gamedesign}). We emphasise that, because model outputs are sampled
(\texttt{temperature=0.7}, no fixed API seed), only the archived replay JSON is
bit-reproducible; a fresh run from the same map seed produces a different game.

\paragraph{Counting launches.} The engine resolves a launch only at the end of
the turn, after both players have played their half-turn; to display the bomb's
effect the viewer captures one extra display frame after resolution, so the
raw replay contains 92 launch-tagged actions for 46 actual model-issued
launches. All statistics in this paper count \emph{model-issued} launches only
(46), unless explicitly noted.

\paragraph{A note on deliberation time and providers.} Because models are
served through three different providers, the measured per-turn think time
reflects not only the model's internal reasoning but also provider-side
scheduling, queuing and rate-limiting. The effect is minor relative to the
order-of-magnitude spread between models, but it is non-zero: some providers
throttle long high-effort calls, and a few models (notably MiniMax M3 and MiMo
2.5 Pro) accumulate timeout retries partly because of provider latency rather
than pure model deliberation. We therefore treat think time as an indicative,
not exact, measure of reasoning effort. We rely primarily on the
illegal-action rate for cross-model comparison, which is provider-independent.
Tokens/turn is only \emph{partially} model-intrinsic: it is confounded by the
fact that different providers expose (or strip) reasoning tokens differently,
so the reasoning-token component of tokens/turn is not strictly comparable
across providers; we report it nonetheless as a coarse indicator, and the
correlations involving tokens/turn (Section~\ref{sec:results}) should be read
with this caveat.

\section{Results}
\label{sec:results}

\FloatBarrier

\subsection{Victory channels}
\begin{figure}[!htbp]
\centering
\includegraphics[width=\linewidth]{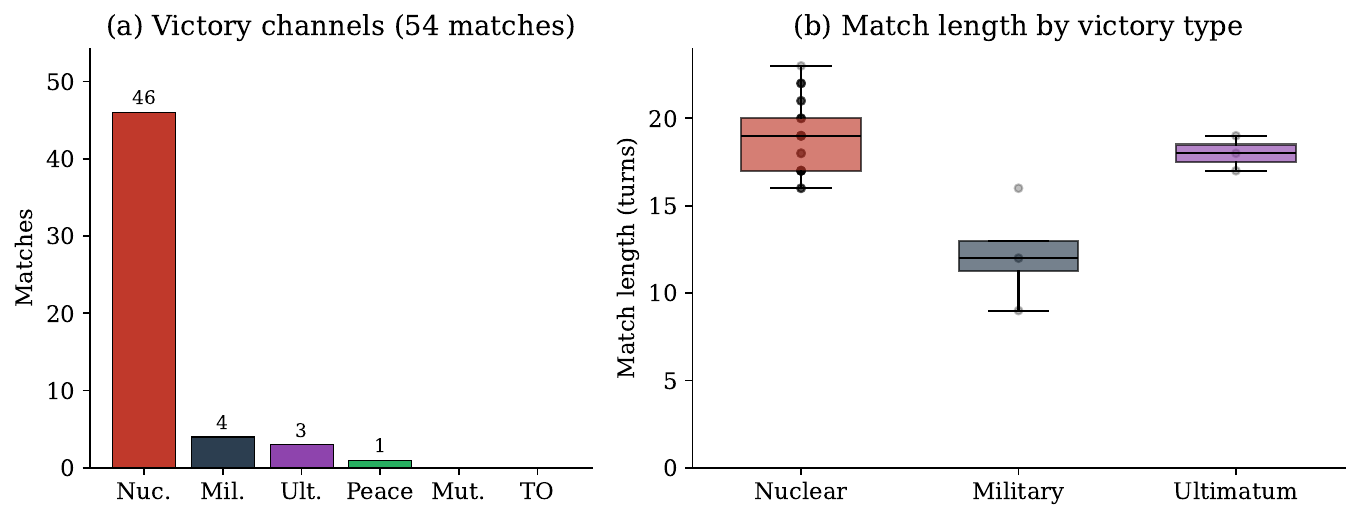}
\caption{(a) Victory channels across 54 completed matches. The nuclear rush
accounts for 78\% of outcomes on the rules-coherent v0.11+ sub-corpus (85\%
corpus-wide, see text); peace occurs once, mutual destruction and timeout
never. (b) Match length by victory type. Military conquests, though rare, end
the match substantially faster (mean 12.3 turns) than nuclear wins (18.9).}
\label{fig:victory}
\end{figure}

Across the 54 completed matches, the corpus-wide outcome distribution is:
\textbf{nuclear 46 (85\%)}, \textbf{military 4 (7\%)}, \textbf{ultimatum 3 (6\%)},
\textbf{peace 1 (2\%)}, mutual destruction 0, timeout 0 (Figure~\ref{fig:victory}a).
Because this mix pools 7 engine versions whose rules change the
military/nuclear balance, our \emph{primary} estimate is the rules-coherent
v0.11+ sub-corpus reported immediately below (nuclear 78\%, military 11\%);
we keep the corpus-wide figures here for continuity with the public
leaderboard and because the qualitative findings (military wins faster,
diplomacy rare) are stable across both cuts. Match
length averages 18.3 turns (range 9--23). The nuclear path is overwhelmingly
dominant; the military conquest path, which the v0.11 rebalance (base HP
8$\to$4) was specifically intended to make competitive, is realised by only
three models (GPT-5.5 once, GLM 5.1 twice, Kimi K2.7 Code once). Strikingly,
military wins are \emph{faster}: mean 12.3 turns (min 9, max 16) versus 18.9 for
nuclear wins (min 16, max 23, Figure~\ref{fig:victory}b). The fastest victory
in the entire corpus is GLM 5.1's 9-turn tank conquest of Grok 4.3; when a
model does commit to the ground push, it ends the game earlier than the nuclear
cycle, suggesting the military path is high-skill but under-attempted rather
than inherently slow. The lone peace was accepted in the GLM~5.2 vs.\ DeepSeek
V4 Pro match (Section~\ref{sec:diplo}).

\paragraph{Primary estimate: the rules-coherent v0.11+ sub-corpus.} The
headline outcome mix is computed on the \textbf{36 matches played under
v0.11.0--0.15.0}, the rules-coherent sub-corpus in which the
military-determining mechanics (base HP 4, line-of-sight blocking, resource
depletion, enemy-deposit reveal) are all fixed. This gives
\textbf{nuclear 28 (78\%)}, \textbf{military 4 (11\%)},
\textbf{ultimatum 3 (8\%)}, \textbf{peace 1 (3\%)}. We treat 78\% as the
cleanest available estimate of nuclear dominance under the described rules.
The full 54-match corpus (which adds 18 pre-v0.11 matches played under an
8-HP base, no LOS blocking and no depletion---all resolved exclusively by
nuclear victory) gives a more lopsided \textbf{85\% nuclear / 7\% military};
the gap (78\% vs.\ 85\%, 11\% vs.\ 7\%) confirms that the pre-v0.11 rules
made the military path structurally harder, inflating the corpus-wide nuclear
share. We therefore lead with 78\% and report 85\% only as the
corpus-wide robustness figure. The illegal-action rate is stable across the
two cuts (5.9\% v0.11+ vs.\ 5.6\% full corpus).

We are explicit that v0.11+ is \emph{not} a single-ruleset corpus: it still
spans v0.11.0 (17 matches), v0.12.0 (8), v0.14.0 (9) and v0.15.0 (2), with a
minor SAM-move-range / mine-cost rebalance introduced in v0.12.0. No single
version has enough matches to serve as the corpus (v0.15.0 has only 2), so
v0.11+ is the best compromise available; the residual inter-version drift is
minor relative to the 8-HP-vs-4-HP change that defines the pre/post-v0.11
split. A single-ruleset study on a future larger v0.15.0 corpus would tighten
the estimate further.

\FloatBarrier

\subsection{The nuclear signature: sole-launcher, single fire}
\begin{figure}[!htbp]
\centering
\includegraphics[width=0.82\linewidth]{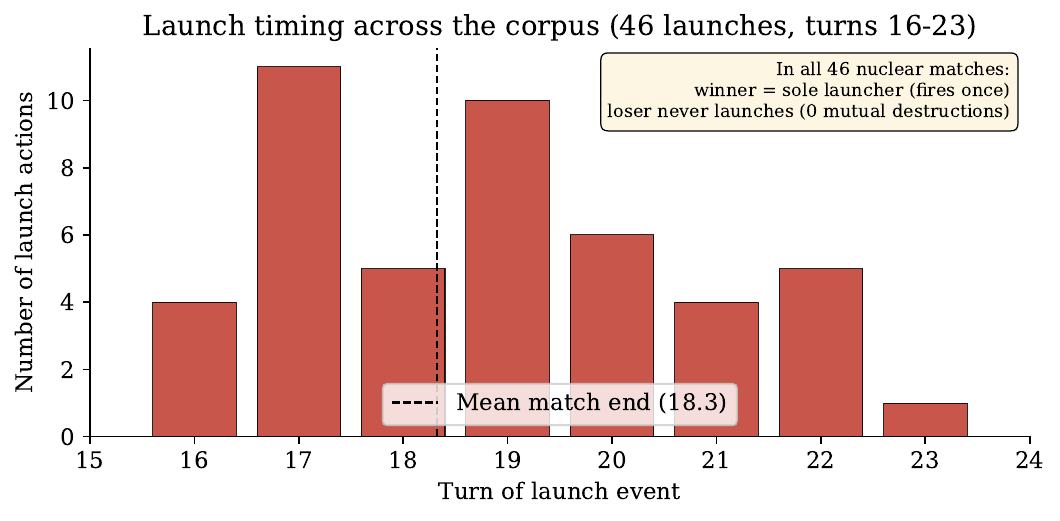}
\caption{Launch timing across the corpus. All 46 model-issued launches fall in
turns 16--23, clustering around the mean match end (18.3 turns). The inset
summary states the signature regularity: in every nuclear match the winner was
the sole launcher, the loser never launched, and no mutual destruction
occurred---a pattern that is largely mechanical under the secret-simultaneous
launch rules (see text), not a cognitive deterrence failure.}
\label{fig:launch}
\end{figure}

A fine-grained inspection of the launch events reveals a stark behavioural
regularity (Figure~\ref{fig:launch}). In \emph{all 46} nuclear matches, the
eventual winner was the \emph{only} player to ever issue a launch, and it did
so exactly once, in the decisive turn; the loser launched \emph{zero} times
across the entire match. Consequently no match ever featured two distinct
players launching in the same turn, and the mutual-destruction outcome, designed
into the engine as the symmetric equilibrium of simultaneous launches, was
\emph{never} observed. Launches occur exclusively in turns 16--23 (peaking at
turn 17, 11 matches), i.e.\ launches cluster tightly around the average match
end (turn 18.3) and \emph{no} model achieves an early
rush.

A silo is \emph{not} consumed by a launch: the
building survives the firing, so a silo whose launch was rejected (e.g.\ for
insufficient uranium) could retry on a later turn; but since a successful
launch ends the match, no winner ever fired more than once. Separately, a silo
is a destructible building (it has HP and can be killed by enemy tanks); across
the corpus a silo is built 112 times over 54 matches, i.e.\ silos are
frequently destroyed and rebuilt within a match, but this never prevented a
nuclear outcome because the winner's silo was always
operational at the decisive turn.

We read the signature with care, because under the rules it is \emph{largely
mechanical, not cognitive}. A launch resolves only at the end of the turn,
after both players have acted; the in-flight bomb is secret (uranium is
hidden, and there is no launch cue except the v0.15.0 early-warning signal,
which fires only if the player currently sees an enemy silo and \emph{never
fired} in any replay). In \emph{all 46 nuclear matches} the eventual loser
therefore had \emph{no information} that a launch was in flight at the moment
it had to decide whether to counter-launch (we restrict the claim to the 46
nuclear matches, since the 8 non-nuclear outcomes---military, ultimatum,
peace---involve no launch at all and are irrelevant to deterrence).
Counter-launching in the same turn would have
required the loser to have independently reached bomb readiness
(operational silo + sufficient uranium + scouted enemy base) on exactly that
turn---i.e.\ to be at parity by chance, with no cue. The absence of mutual
destruction is thus primarily a consequence of the secret, simultaneous
resolution design, not evidence that the models ``failed to deter.'' The
genuine, weaker cognitive finding is the absence of \emph{pre-emptive}
launches: no losing model, even when it had built a silo, launched
speculatively to guard against a possible enemy launch. Whether the
v0.15.0 early-warning signal (when active) would induce counter-launches
and mutual destructions is an open question, and a natural next experiment.

\FloatBarrier

\subsection{Leaderboard (preliminary descriptive view)}
\label{sec:leaderboard}
We present the leaderboard as a \emph{preliminary descriptive view}, not a
contribution of the paper. Match counts per model are uneven (3--15), the
pairing is not a balanced round-robin and is not side-swapped
(Section~\ref{sec:setup}), and each match is a single stochastic run
(\texttt{temperature=0.7}, no fixed seed), so there is no run-to-run variance
estimate and a single match can turn on a stochastic token. At this $n$,
points-per-match (ppm) is \emph{not directly comparable} across models that
faced different opponent mixes, and the ordering is indicative only.

Table~\ref{tab:board} ranks models by ppm. GPT-5.5 is undefeated
(7W--0L, 3.00 ppm), but with only 7 matches the undefeated claim is wide on a
Clopper--Pearson win-rate interval ($\approx$[0.59, 1.00]) and should not be
read as established invincibility. The mid-tier is dense (Gemini Flash,
MiniMax M3, Claude Fable, DeepSeek V4 Pro, GLM 5.1, Claude Opus 4.8, GLM 5.2);
their ppm bootstrap CIs overlap almost entirely (e.g.\ Gemini Flash [1.50,
3.00], DeepSeek V4 Pro [1.13, 2.60], GLM 5.2 [0.50, 2.67]), so their relative
ordering is \emph{not} statistically distinguishable at this sample size. The
bottom is occupied by models with either weak reasoning (Grok 4.3, 0W) or
severe reliability issues (Kimi variants, MiMo 2.5 Pro).

\paragraph{An opponent-adjusted ranking (Bradley--Terry).} Because ppm ignores
\emph{who} a model played, we also fit a Bradley--Terry model (each match
contributes a win/loss/draw between two model strengths; draws score 0.5) with
1{,}000-resample bootstrap confidence intervals, which is the standard tool
for unbalanced tournament data. The BT ordering agrees with the ppm
ordering (Spearman $\rho\approx 0.99$ between the two rankings): GPT-5.5,
Gemini Flash 3.5 and MiniMax M3 are the top three in both, and Grok 4.3 /
Kimi 2.6 are last. We caution against reading this agreement as a validation
of robustness: with such unbalanced, low-$n$ data, ppm and BT converge almost
by construction (both are monotone functions of the same sparse win/loss
matrix), so the agreement is expected rather than reassuring. The BT
intervals are very wide (the top-three CIs all overlap), so even the
opponent-adjusted ranking cannot separate the leaders. We report ppm for
continuity with the public leaderboard and treat BT as a sanity check on
gross distortion, not as evidence of a stable ranking; that will require
side-swapped $\geq$20 matches per pairing and repeated runs
(Section~\ref{sec:setup}, future work).

\begin{table}[!htbp]
\centering
\caption{Leaderboard (sorted by points/match). \emph{ppm}=points per match,
\emph{W/L/D}=wins/losses/draws, \emph{tt}=avg think time per half-turn (s),
\emph{tok}=tokens/turn, \emph{inv}=illegal-action rate, \emph{\$}=cost/match
(USD). All models are evaluated on the same deterministic engine. The cost
column reflects API spend (input+output tokens at each provider's pricing) and
is reported as a sampling constraint rather than a performance metric: cheaper
models permit more matches per budget, but cost does not track winning
(GPT-5.5 at \$1.45/match and Grok 4.3 at \$0.10/match sit at opposite ends of
the table).}
\label{tab:board}
\small
\begin{tabular}{lrrrrrrrr}
\toprule
Model & W & L & D & ppm & tt(s) & tok & inv & \$ \\
\midrule
GPT-5.5            & 7 & 0 & 0 & 3.00 & 289 &  7473 & 5.6\% & 1.45 \\
Gemini Flash 3.5   & 8 & 2 & 0 & 2.40 &  50 & 15538 & 4.9\% & 1.80 \\
MiniMax M3         & 5 & 2 & 0 & 2.14 & 662 & 15687 & 6.1\% & 0.25 \\
Claude Fable 5     & 2 & 1 & 0 & 2.00 &  55 &  6866 & 2.5\% & 1.91 \\
DeepSeek V4 Pro    & 9 & 5 & 1 & 1.87 & 151 &  6682 & 5.6\% & 0.24 \\
GLM 5.1            & 6 & 4 & 0 & 1.80 & 133 &  8373 & 6.4\% & 0.15 \\
Claude Opus 4.8    & 5 & 4 & 0 & 1.67 &  59 &  6578 & 3.0\% & 1.16 \\
GLM 5.2            & 3 & 2 & 1 & 1.67 & 188 & 14878 & 6.6\% & 0.83 \\
Gemini Pro 3.1     & 2 & 3 & 0 & 1.20 &  53 & 11361 & 1.1\% & 1.39 \\
Nemotron 3 Ultra   & 2 & 6 & 0 & 0.81 &  38 &  9854 & 3.2\% & 0.25 \\
Kimi K2.7 Code     & 1 & 4 & 0 & 0.70 & 323 & 25486 & 7.0\% & 1.20 \\
MiMo 2.5 Pro       & 2 & 7 & 0 & 0.67 & 326 &  7764 & 5.6\% & 0.15 \\
Qwen 3.7 Max       & 1 & 4 & 0 & 0.60 &  75 &  5688 & 6.4\% & 0.29 \\
Kimi 2.6$^\dagger$ & 0 & 5 & 0 & 0.10 & 257 & 28869 & 8.0\% & 1.54 \\
Grok 4.3           & 0 & 4 & 0 & 0.00 &   7 &  5278 & 8.6\% & 0.10 \\
\bottomrule
\end{tabular}\\
\footnotesize $^\dagger$ archived (superseded by K2.7).
\end{table}

\begin{figure}[!htbp]
\centering
\includegraphics[width=\linewidth]{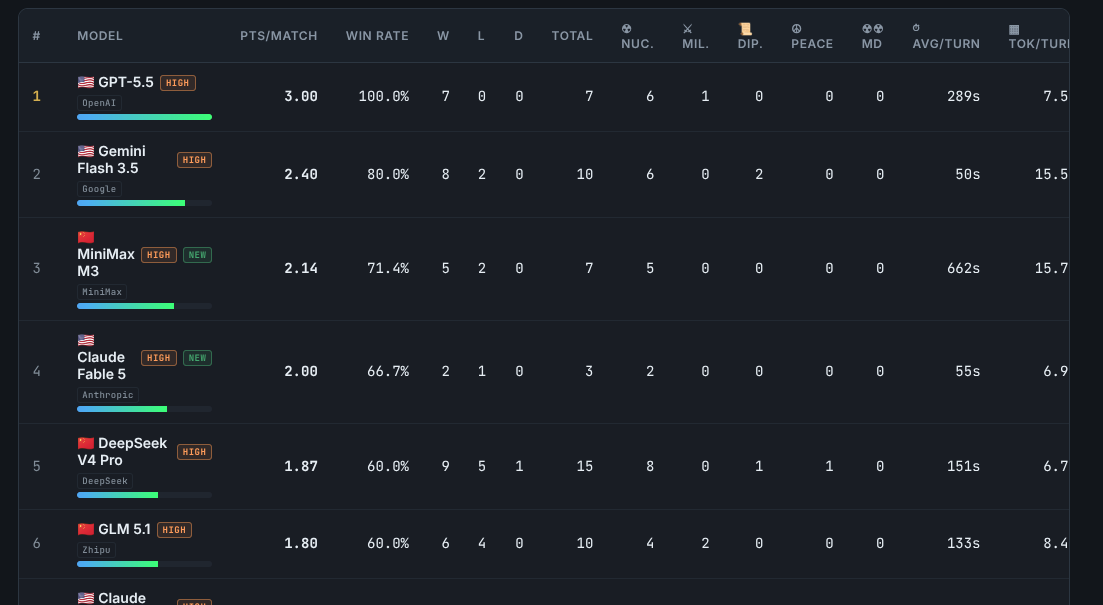}
\caption{The public live leaderboard at \url{https://ageofllm.org}, ranked by
points per match. The frozen 54-match snapshot analysed in this paper (Table~\ref{tab:board})
is a strict subset of this continuously updated view.}
\label{fig:leaderboard_view}
\end{figure}

\FloatBarrier

\subsection{Deliberation does not predict winning}
\begin{figure}[!htbp]
\centering
\includegraphics[width=0.78\linewidth]{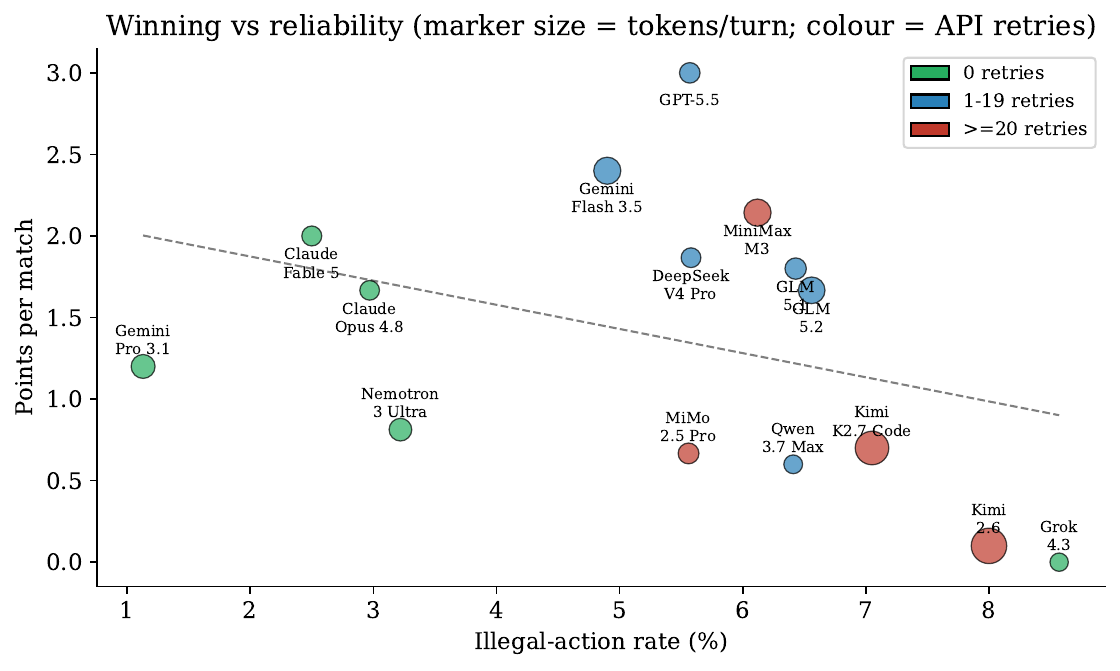}
\caption{Points per match versus illegal-action rate. The negative correlation
($r=-0.35$, $p=0.20$, $n=15$) is the strongest of the three single correlates
of winning we tested, exceeding deliberation time ($r=+0.19$) and tokens/turn
($r=-0.24$, encoded as marker size), but none is individually significant and
the trend line is illustrative only (see text for Steiger test and
Bonferroni-corrected interpretation). Marker colour encodes the model's total
API retries: green $=0$, blue $=$ 1--19, red $\geq 20$. The most error-prone
models (Grok 4.3, Kimi 2.6) sit at the bottom; notably Grok 4.3 records zero
API retries (green) yet still loses because of its high illegal-action rate
(8.6\%), illustrating that action-level reliability, not just API-level
reliability, accompanies losing at the bottom of the board.}
\label{fig:reliability}
\end{figure}

With only $n=15$ models, every correlation in this subsection has low
statistical power and wide uncertainty; we report $p$-values and 95\% bootstrap
confidence intervals (10{,}000 resamples) and treat the patterns as
\emph{exploratory} rather than confirmed. Pearson correlations between
points/match and model-level aggregates are weak and individually
non-significant: illegal-action rate $r=-0.35$ ($p=0.20$, 95\% CI
$[-0.73,+0.17]$; Spearman $\rho=-0.45$, $p=0.09$), deliberation time
$r=+0.19$ ($p=0.51$, CI $[-0.48,+0.57]$; $\rho=+0.13$), and tokens/turn
$r=-0.24$ ($p=0.39$, CI $[-0.69,+0.56]$; $\rho=+0.10$). The illegal-action rate
is nonetheless the strongest (negative) single correlate of winning
(Figure~\ref{fig:reliability}). Because these correlations share the
points/match variable, we compare them with Steiger's $Z$ test for dependent
correlations. We treat the \emph{tokens/turn} comparison as the primary formal
test, because deliberation time is provider-confounded (Section~\ref{sec:setup})
and feeding a noisy variable into the Steiger test would bias it toward the
null and could manufacture a spurious ``borderline'' gap. The gap between the
illegal-action correlation and the tokens/turn correlation is not significant
($p=0.71$). For completeness, the illegal-vs-think gap is borderline
($t=-2.18$, $p=0.0499$), but given the provider confound on think time we do
not interpret this $p=0.0499$ as robust. Correcting for
the five correlations we tested (illegal, think, tokens, courtesy, bluff) with
Bonferroni, \emph{none} remains significant at $\alpha=0.05$; the directional
ordering (illegal $>$ think $\approx$ tokens in predictive strength) is
therefore suggestive, not established. Descriptively, GPT-5.5 thinks nearly
six times longer than Gemini Flash per half-turn yet both sit near the top;
conversely Kimi variants burn the most tokens (25k--29k/turn) and lose. We
read this as tentative evidence that, under this benchmark's sequential,
partially-observable regime, the \emph{action-legality} dimension tracks
winning at least as well as raw deliberation volume, while noting that
action-legality is itself a product of reasoning (belief-tracking), so the two
are not independent. We caution that the deliberation-time comparison is
partly confounded by provider differences (see Section~\ref{sec:setup}): some
models are served through providers that throttle or queue long calls, so the
measured think time is not a pure reflection of model reasoning. The
illegal-action rate, by contrast, is provider-independent and is the metric we
trust most; tokens/turn is only partially model-intrinsic (reasoning-token
exposure varies by provider, Section~\ref{sec:setup}) and is reported as a
coarse indicator only.

\FloatBarrier

\subsection{Diplomacy is proposed but rarely consummated}
\label{sec:diplo}
\begin{figure}[!htbp]
\centering
\includegraphics[width=0.78\linewidth]{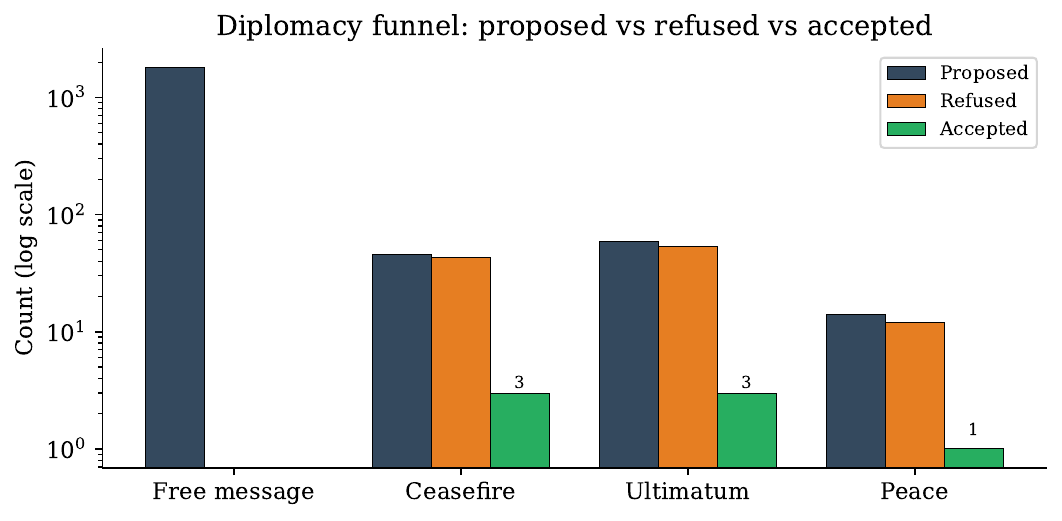}
\caption{Diplomacy funnel (log scale). Free messages dominate traffic (1{,}804)
but carry no binding effect. Of the binding proposals, only ultimatums ever
converted to a victory (3/59); peace was accepted once (1/14, a draw) and
ceasefires only 3/46.}
\label{fig:diplo}
\end{figure}

Across the corpus we count 1{,}804 free messages, 46 ceasefire proposals, 59
ultimatums and 14 peace proposals (Figure~\ref{fig:diplo}). Yet only \emph{one}
of the 14 peace proposals was accepted (12 explicitly refused, 1 ignored) and
only 3 of 46 ceasefires were accepted (43 refused, 0 ignored, every
ceasefire drew a response). The sole
diplomatic path to victory was the ultimatum: 3 of 59 were accepted (53
refused, 3 ignored), and those three acceptances are exactly the three
ultimatum victories (two by Gemini Flash 3.5 and one by DeepSeek V4 Pro).

The two binding settlement channels have structurally different incentives,
and the data reflects this asymmetry. \textbf{Peace} is terminal: once
accepted, the match ends \emph{immediately} as a draw (1 point each), so the
opponent \emph{cannot} launch a bomb afterwards. Refusing peace is therefore
not a safety concern but a value bet: a model refuses peace when it believes it
can still win (3 points beats 1), which is rational whenever its nuclear or
military push is on track. The single accepted peace occurred in the
GLM~5.2 vs.\ DeepSeek V4 Pro match at turn 16, and it is the one match where
the message channel produced a non-nuclear outcome. Both sides had scouted the
other's base early (by turn 5) and both had built a silo, but they were far
apart in the uranium race: DeepSeek held 24U (one short of the 25U bomb cost)
with an operational silo, while GLM~5.2 lagged at 11U and could not launch.
DeepSeek used the message channel to issue an explicit launch threat
(``I have 25 uranium now and my silo is operational; accept peace for a draw, or
face defeat''), and GLM~5.2, judging the threat credible and unable to
counter-launch in time, accepted the guaranteed draw over likely nuclear defeat
(``Your uranium position was credible and I couldn't reach your silo in time. A
draw is better than mutual destruction''). This is the corpus's sole instance
of successful deterrence via communication, and it is telling that it took a
model that was genuinely behind (GLM~5.2) to prefer 1 guaranteed point over a
contested 3. The near-absence of accepted peace elsewhere is consistent with
models correctly assessing that, in a race the average match resolves by turn
18, neither side is usually firmly enough behind to prefer a guaranteed 1 point
over a contested 3.

\textbf{Ceasefire}, by contrast, is the channel that carries a genuine
strategic dilemma. A ceasefire forbids conventional \emph{attacks} for three
turns, but it does \emph{not} stop the nuclear bomb: either side may still
launch during a ceasefire at a +6U penalty. Accepting a ceasefire therefore
grants the opponent up to three uninterrupted turns to finalise its silo and
uranium stockpile and launch in relative safety, while denying you the ability
to disrupt it with tanks. The low ceasefire acceptance rate (3/46) suggests
models intuit this risk, even implicitly. The ultimatum channel works precisely
because accepting it is strictly better than a clean defeat (0.5 vs.\ 0), so a
genuinely losing model has a dominant incentive to accept only \emph{that}
channel.

\FloatBarrier

\subsection{Action budget and combat}
\begin{figure}[!htbp]
\centering
\includegraphics[width=\linewidth]{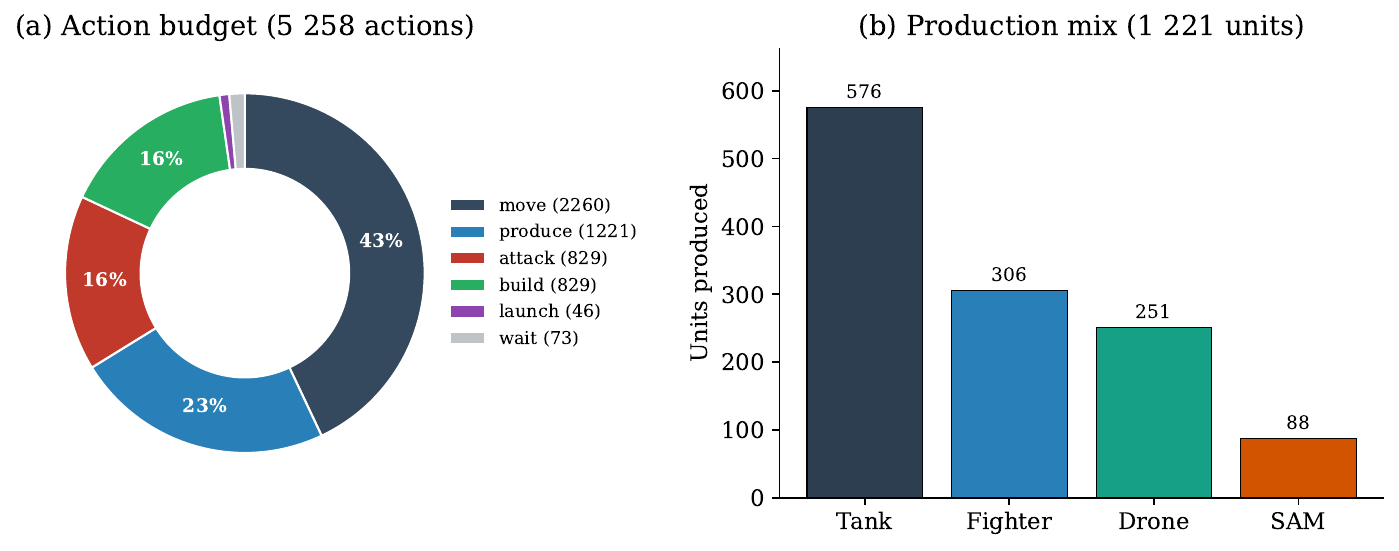}
\caption{(a) Action budget across the corpus (5{,}258 actions). Movement
dominates (43\%); launches are only 1\% of all actions yet decide 78\% of
matches on the rules-coherent sub-corpus (85\% corpus-wide). (b) Production mix (1{,}221 units). Tanks are the most produced unit
yet rarely achieve the base-conquest win condition.}
\label{fig:actions}
\end{figure}

The 54 matches comprise 5{,}258 emitted actions, distributed as in
Figure~\ref{fig:actions}: move 2{,}260 (43\%), produce 1{,}221 (23\%), attack
829 (16\%), build 829 (16\%), launch 46 (1\%) and wait 73 (1\%) --- the equality
of the attack and build totals (829 each) is coincidental; they are counted
from disjoint action types and the build total cross-checks against its
sub-components (411+166+140+112 = 829).
Production is
tank-heavy (tanks 576, fighters 306, drones 251, SAMs 88); yet only 4 matches
ended in a tank conquest of the base, so tanks are overwhelmingly used for
centre-contestation and defence rather than base assault. Building is
economy-led: 411 credit mines, 166 uranium mines, 140 central uranium mines and
112 silos; yet only 46 of these ever fired, one per nuclear winner, so 66 of
the 112 silo-builds (59\%) never fired at all (a silo may be rebuilt on the
same cell after destruction, so these are build events, not distinct
buildings), either destroyed by enemy tanks or rendered
moot when the match ended by another route. A silo is
a destructible building with HP, and across the corpus silos are rebuilt within
a match after being destroyed; the engine does not forbid rebuilding on a freed
cell, so silo turnover is common but never blocked a nuclear outcome. Combat is highly lethal: of 829 attacks, 730 (88\%) destroyed
their target (a unit or building) and 99 (12\%) merely damaged a building, a
consequence of the HP-less unit design where every valid hit on a unit is
fatal.

\FloatBarrier

\subsection{What illegal actions reveal about partial observability}
\begin{figure}[!htbp]
\centering
\includegraphics[width=0.85\linewidth]{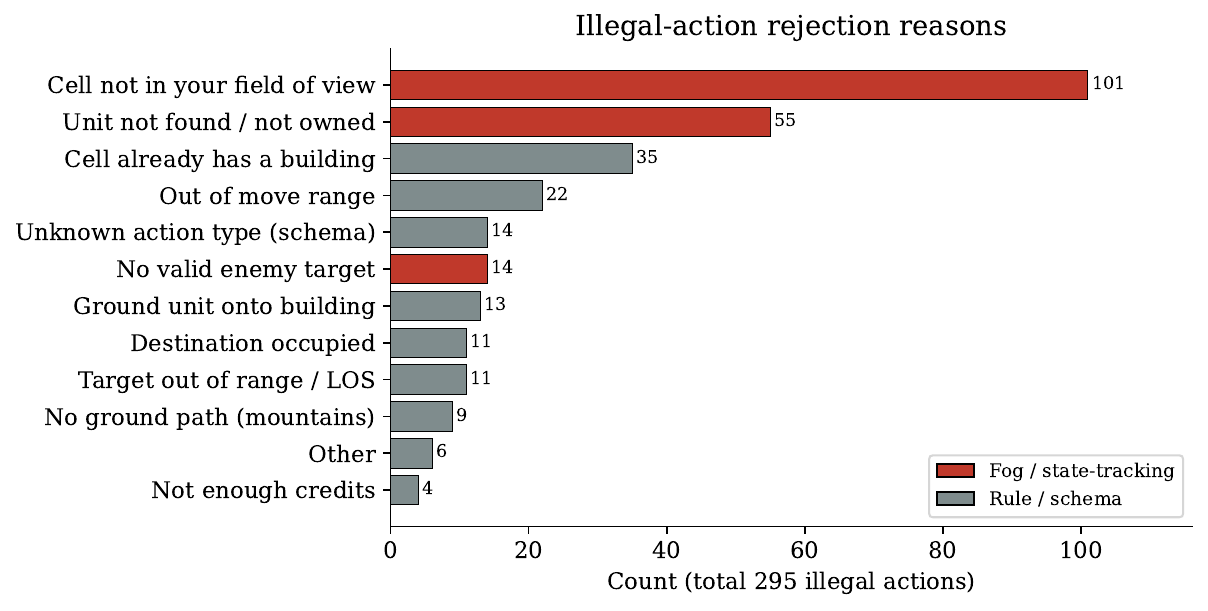}
\caption{Illegal-action rejection reasons (295 total). Red bars are fog/state
errors (field-of-view and stale/imagined unit references); grey bars are
rule/schema errors. Fog and state tracking account for $\sim$53\% of all
illegal actions.}
\label{fig:illegal}
\end{figure}

Of the 5{,}258 actions, 295 (5.6\%) were rejected by the engine as illegal.
Categorising the rejection reasons (Figure~\ref{fig:illegal},
Table~\ref{tab:invreason}) shows that the two largest classes are \emph{not}
rule misunderstandings but direct consequences of fog of war and state
tracking: ``cell not in your field of view'' accounts for 101 (34\%) and
``unit not found / not owned'' for 55 (19\%), the latter reflecting references
to stale or imagined units. Together, these two largest classes are $\sim$53\%
of all illegal actions; a third, smaller state-tracking class, ``no valid
enemy target at cell'' (14, i.e.\ attacking a cell where the believed enemy no
longer is), is also a belief failure and is coloured as fog/state in
Figure~\ref{fig:illegal}, raising the full fog/state share to $\sim$58\%.
Rule-proper errors (line-of-sight blocked, move-range
exceeded, occupancy, insufficient credits) make up the remainder. This
reframes the illegal-action rate less as a measure of instruction-following and
more as a measure of \emph{belief-tracking under partial observability},
precisely the competence the benchmark is designed to probe.

\begin{table}[!htbp]
\centering
\caption{Top rejection reasons for the 295 illegal actions. Fog/state errors
(field-of-view and stale/imagined unit) dominate.}
\label{tab:invreason}
\small
\begin{tabular}{lrr}
\toprule
Rejection reason & count & \% \\
\midrule
Cell not in your field of view (scout first) & 101 & 34.2 \\
Unit not found / not owned                    &  55 & 18.6 \\
Cell already has a building                   &  35 & 11.9 \\
Out of move range                             &  22 &  7.5 \\
No valid enemy target at cell                 &  14 &  4.7 \\
Unknown action type (schema error)            &  14 &  4.7 \\
Ground unit cannot move onto a building       &  13 &  4.4 \\
Target out of range / line-of-sight blocked   &  11 &  3.7 \\
Destination occupied (ground/air unit)        &  11 &  3.7 \\
No ground path (mountains block)               &  9 &  3.1 \\
Not enough credits                             &  3 &  1.0 \\
Other                                         &  7 &  2.4 \\
\bottomrule
\end{tabular}
\end{table}

\FloatBarrier

\subsection{Reliability as a discriminator}
Table~\ref{tab:retries} shows API retry counts by cause (a retry is any failed
API attempt, i.e.\ a timeout, transport error, or unparseable JSON, within the
\texttt{max\_retries}=3 budget of a single half-turn). Five models (Claude
Opus/Fable, Gemini Pro 3.1, Nemotron 3 Ultra, Grok 4.3) record zero retries,
while the three losing retry-heavy models---MiMo 2.5 Pro, Kimi 2.6 and Kimi
K2.7 Code---each accumulate 24--36 retries. MiniMax M3 is a notable
\emph{exception}: it records 28 retries (timeout-dominated, partly
provider-latency, see Section~\ref{sec:setup}) yet still ranks near the top
(2.14 ppm), showing that high retry counts do not by themselves doom a model
that otherwise emits legal, effective actions.
MiMo 2.5 Pro spans all three buckets (timeouts, API errors and malformed JSON),
whereas the Kimi variants are malformed-JSON dominated (24 malformed each,
little else). Crucially,
a failed half-turn does \emph{not} silently become a \texttt{wait}: when all
three attempts fail, the engine \emph{pauses} the match for a cool-down period
and then re-runs the same half-turn from scratch (a deliberate wait chosen by the
model exits immediately and is never confused with a failure). These pauses are
not captured in the replay, so the retry counts below reflect in-half-turn
instability only. Across the 108 half-turns that incurred at least one retry,
only 2 (both MiMo 2.5 Pro) exhausted all three attempts and triggered a
pause-and-resume; the rest recovered on a later attempt and lost no turn. The
retry count is therefore a \emph{proxy} for inference instability rather than a
direct count of lost turns; nonetheless, models that retry heavily also tend to
emit more illegal actions and lose more often, so the signal is consistent.

\begin{table}[!htbp]
\centering
\caption{API reliability per model (total retries across all its matches,
bucketed by cause).}
\label{tab:retries}
\small
\begin{tabular}{lrrrr}
\toprule
Model & timeout & api\_err & malformed & total \\
\midrule
MiMo 2.5 Pro    & 14 & 8  & 14 & 36 \\
Kimi 2.6        &  1 & 0  & 24 & 25 \\
MiniMax M3      & 19 & 3  &  6 & 28 \\
Kimi K2.7 Code  &  0 & 0  & 24 & 24 \\
GLM 5.2         &  0 & 2  &  6 &  8 \\
GPT-5.5         &  4 & 1  &  2 &  7 \\
DeepSeek V4 Pro &  1 & 4  &  0 &  5 \\
Gemini Flash 3.5&  0 & 0  &  2 &  2 \\
GLM 5.1         &  0 & 0  &  1 &  1 \\
Qwen 3.7 Max    &  0 & 1  &  0 &  1 \\
Claude Opus 4.8 & 0 & 0  &  0 &  0 \\
Claude Fable 5  & 0 & 0  &  0 &  0 \\
Gemini Pro 3.1  & 0 & 0  &  0 &  0 \\
Nemotron 3 Ultra& 0 & 0  &  0 &  0 \\
Grok 4.3        & 0 & 0  &  0 &  0 \\
\bottomrule
\end{tabular}
\end{table}

\FloatBarrier

\subsection{Emergent communication: courtesy, bluff and the missing GG}
The free-message channel (1{,}804 messages across the corpus) carries no
binding effect, yet models use it in qualitatively distinct ways that reveal
emergent ``personality'' and, notably, \emph{deception}. We split the 1{,}804
messages into opening (turn $\leq$ 3, $n=320$), midgame ($n=1{,}128$) and endgame
(last 4 turns, $n=356$).

\begin{figure}[!htbp]
\centering
\includegraphics[width=\linewidth]{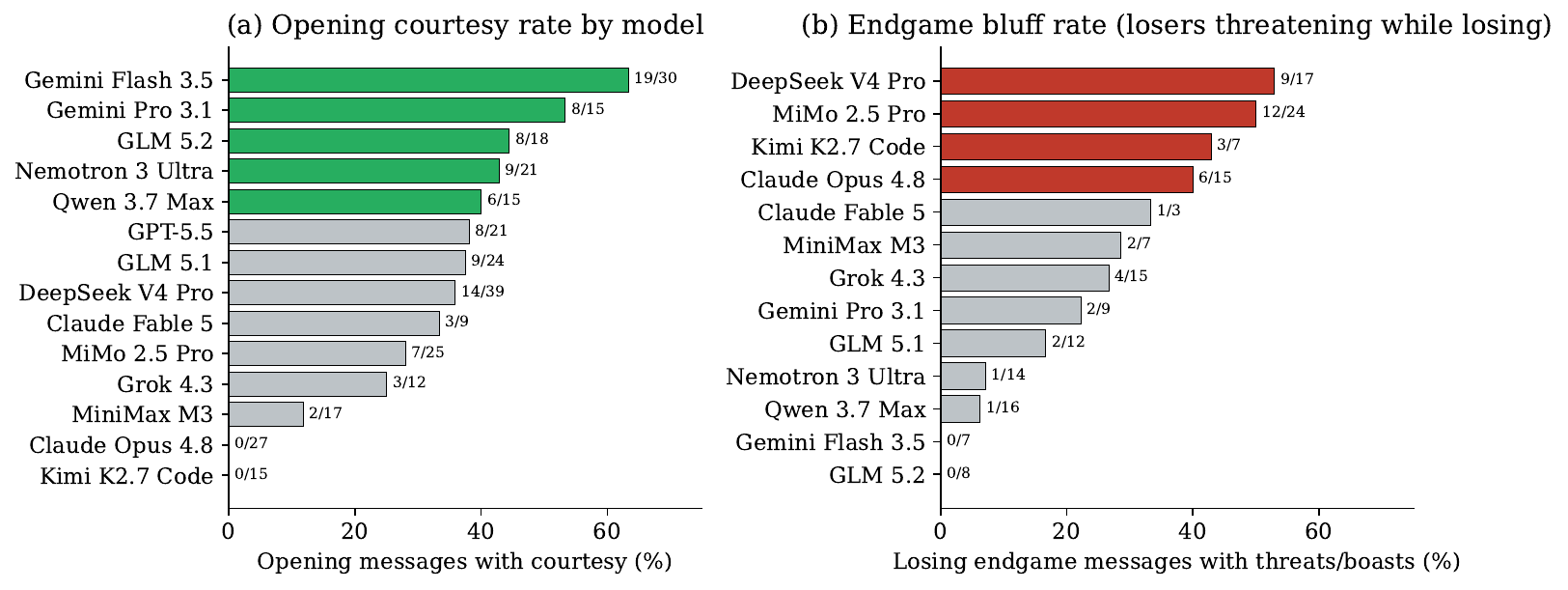}
\caption{(a) Opening courtesy rate by model. Courtesy phrases appear in up to
63\% of Gemini Flash openings but never in Claude Opus or Kimi K2.7; the rate
is weakly and non-significantly correlated with points/match ($r=+0.35$,
$p=0.20$, $n=15$), marking a stylistic trait that co-occurs with, rather than
causes, strong play. (b) Endgame bluff rate among losers (nuclear and military
matches only, where the winner/loser split is unambiguous): the share of a
losing model's final-turn messages that contain threats or boasts. Roughly
28\% of losing endgame messages bluff, peaking near 50\% for MiMo 2.5 Pro and
DeepSeek V4 Pro, evidencing emergent deception despite the rule-only prompt
(classifier-based; see Section~\ref{sec:results} message-coding method).}
\label{fig:comm}
\end{figure}

\paragraph{Message-coding method.} All message categories in this and the
following subsection are produced by a \emph{deterministic, lexicon-based
classifier} rather than human annotation. Each message is lowercased and tested
for substring matches against fixed, hand-curated keyword lists:
\texttt{courtesy} = \{\emph{good luck, may the best, best general, best
commander, best strategist, great game, good game}\}; \texttt{threat/boast} =
\{\emph{surrender, crush, destroy, nuclear, launch, bomb, nuke, strike, finish,
end this, defeat, you will, annihilat, overrun, roll}\}; and a
\texttt{placid/economy} list for the deceptive-calm analysis
(Section~\ref{sec:secrecy}). A message is counted in a category if it contains
any list keyword; the endgame bluff counts (Fig.~\ref{fig:comm}b) are
restricted to nuclear and military victories, where the winner/loser split is
unambiguous, so the 168/168 loser/winner totals there exclude the 20 endgame
messages from the 3 ultimatum and 1 peace matches (which have no such split,
leaving $356 - 168 - 168 = 20$). Because the classifier is deterministic and
the keyword lists are published alongside the analysis scripts, these counts
are exactly reproducible; the trade-off is the usual one for lexicon methods
(no inter-annotator agreement statistic, sensitivity to synonymy and
polysemy, and a bias toward under-counting indirect phrasings). We therefore
treat the rates as descriptive indicators of message tone rather than as
ground-truth pragmatic labels.

\paragraph{Opening courtesy is a stable stylistic trait.} The fraction of
opening messages containing a courtesy phrase (``good luck'', ``may the best
strategist win'', ``great game'') varies sharply by model and is weakly
correlated with skill ($r=+0.35$, $p=0.20$ against points/match; exploratory).
Gemini Flash 3.5 and Gemini Pro 3.1 open courteously in 63\% and 53\% of their
opening messages
respectively, whereas Claude Opus 4.8 (0\%, across 27 opening messages) and Kimi
K2.7 Code (0\%, across 15) never do, instead reporting tactical facts
(``scouting the center and securing our economy''). Courtesy is therefore
best read as a learned persona that co-occurs with, rather than causes,
strong play (Figure~\ref{fig:comm}a).

\paragraph{Endgame bluff is common and symmetric.} Of 168 endgame messages
sent by the eventual \emph{loser}, 47 (28\%) contain a threat or boast
(``surrender or be crushed'', ``bring your bomb, we'll make sure you don't live
to see the fallout'', ``I'm not done yet'') despite the model being on a losing
trajectory. Winners boast at a comparable rate (46/168, 27\%), so the message
tone alone is a poor predictor of the true board state. The symmetry is
exact, not merely descriptive: a $\chi^2$ test on the $2\times2$
(loser/winner $\times$ threat/no-threat) gives $p=0.90$ (uncorrected; Fisher
exact $p>0.99$), and a per-match paired
$t$-test on each match's loser-vs-winner bluff rate gives $t=0.00$, $p=1.0$
($n=50$ matches with messages on both sides; the per-match differences are
non-zero in 30/50 matches but cancel exactly in aggregate). We can therefore state that
losers and winners bluff at statistically indistinguishable rates---bluff is
genuinely symmetric, not just apparently so. Bluffing is unevenly
distributed: MiMo 2.5 Pro and DeepSeek V4 Pro bluff in roughly half of their
losing endgame messages, while Gemini Flash 3.5 and Qwen 3.7 Max almost never
do (Figure~\ref{fig:comm}b). This is genuine emergent deception: the models
are not instructed to bluff, the prompt only describes the rules, yet losing
models systematically project strength they do not have.

\paragraph{The missing GG, with an observability caveat.} Despite 93 messages
sent on the decisive final turn, only 5 contain a courtesy/concession marker
(``gg'', ``well played'', ``concede''). Models do not gracefully acknowledge
defeat: they keep threatening, reporting, or boasting up to the frame the
bomb detonates. We caveat this reading with an observability confound that
partly explains it for nuclear matches: because a launch resolves secretly at
turn end, the eventual loser did not know it had lost while it still had a turn
to message (Section~\ref{sec:results}), so the absence of a concession is
partly an artefact of information, not purely a failure of social-convention
transfer. The cleaner test is the 4 military victories, where the loser
watches its base fall to tanks on-screen and therefore \emph{does} know it is
losing while it can still message: even there, concession markers remain
absent (0 of the military-defeat final-turn messages contain one), though the
$n=4$ is too small to draw a firm conclusion. The finding we can defend is
weaker than ``models fail to concede'': it is that competitive vocabulary is
imported (threats, boasts, courtesy openers) but closing etiquette is not, and
the nuclear-resolution mechanic makes the cleanest version of this test
infeasible at the current $n$.

\FloatBarrier

\subsection{Deceptive calm: hiding the silo and the launch}
\label{sec:secrecy}
\begin{figure}[!htbp]
\centering
\includegraphics[width=0.85\linewidth]{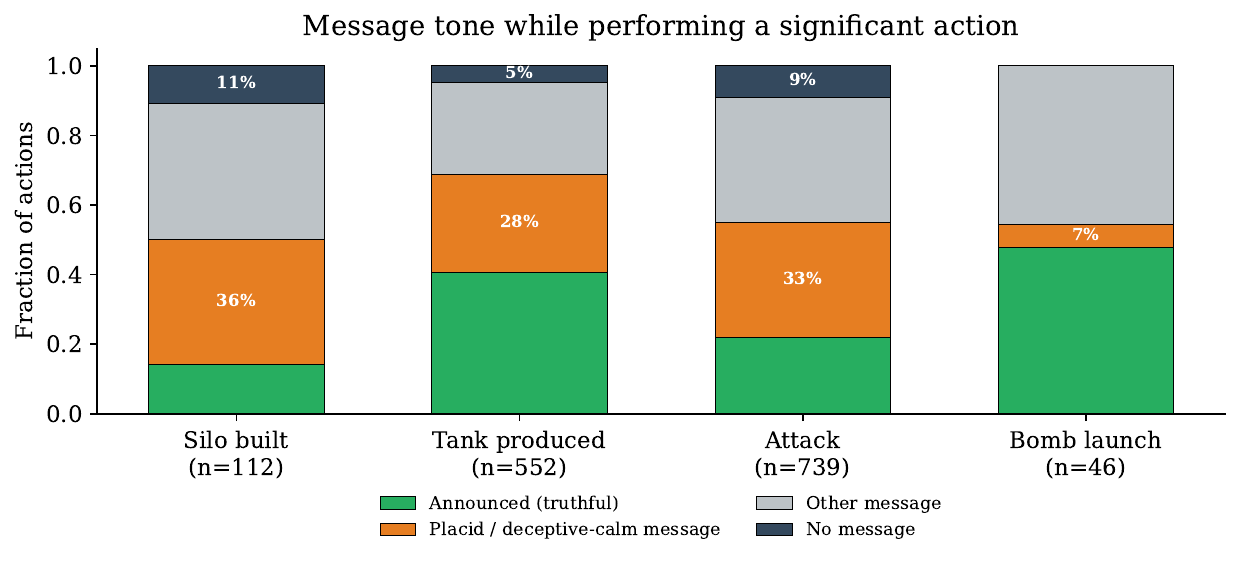}
\caption{Message tone while performing a significant action. Green = the
message truthfully announces the action; orange = a placid/economy message that
masks it; grey = other message; dark = no message at all. Silo construction is
the least announced action (only 14\%), markedly below comparably mundane
actions (tank production 41\%, attacks 22\%); the launch itself is announced
48\%, a figure partly confounded by turn salience (see text).}
\label{fig:secrecy}
\end{figure}

A finer analysis aligns each message with the actions the sender \emph{actually
performed that same half-turn}, revealing a second mode of emergent deception
that is not a false boast but a \emph{false calm} (Figure~\ref{fig:secrecy}).
Since each half-turn carries a single optional message, the fractions below are
computed per half-turn in which the action was taken (the $n$ shown in the
figure), not per individual action. When a player builds a silo, only 14\% of
the time does the accompanying message acknowledge anything nuclear; in 36\% of
cases the message is placid or economy-flavoured (``securing our resource
foundation'', ``peaceful development continues''), i.e.\ the player conceals a
strategic-weapon build behind banal economic chatter. Attacks are similarly
masked: in turns where a player attacks, only 22\% are announced, while 33\%
arrive wrapped in a placid message and 9\% with no message at all. The launch itself is the most openly declared
high-stakes action: of the 46 model-issued launches, 48\% are announced
outright and 0\% are sent silently, so game-ending bombs are, if anything,
accompanied by a message more often than routine attacks. By contrast, the
\emph{preparation} for a launch, the silo build, is the most concealed action
(only 14\% announced). The silo-vs-launch announce gap is statistically
significant and not merely descriptive: a $\chi^2$ test on
(silo/launch $\times$ announced/silent) gives $p=2\times10^{-5}$ (Fisher
exact $p=2\times10^{-5}$, odds ratio 0.18), and a per-match Mann--Whitney
test on each match's announce rate gives $p=0.001$. The association between
action type and announce rate is therefore established, not just observed.
We are, however, cautious about the \emph{strategic} attribution. A launch is
the terminal, unique action of the decisive turn, so the model often has
nothing else to narrate and a non-trivial fraction of the 48\% may reflect a
general propensity to narrate salient endgame turns rather than a deliberate
decision to announce the bomb; the 0\% silent launches is likewise consistent
with this turn-salience confound. The silo-vs-launch comparison is thus
\emph{confounded} by turn salience.

The cleaner contrast is silo \emph{vs.\ other routine actions}: tank
production (41\% announced) and attacks (22\%) are, like the silo build,
ordinary ``one action among three in a mundane turn,'' yet they are announced
at 2--3$\times$ the silo rate. Turn salience alone cannot explain why the silo
build is singled out for concealment among equally mundane actions, so the
silo-vs-tank/attack gap is the part of the pattern most consistent with a
concealment reading. Tank production sits in between (41\% announced, only
5\% silent), consistent with tanks being a visible, contested centre-piece
rather than a secret weapon.

It is worth noting that the message channel and the action channel are
mechanically independent in the engine: a model may emit a diplomatic message
and a \texttt{launch} action in the same half-turn without restriction, and the
engine does not force, suppress or pair them. Whether a launcher announces its
bomb is therefore a model choice, not an engine constraint.

We therefore phrase the finding conservatively: the data are
\emph{consistent with} concealment of the silo build relative to comparably
mundane actions, but this design cannot separate a concealment motive from
turn-salience (for the launch) or from forgetfulness (for the silo build). The
forgetfulness reading is non-trivial: a model spending its three actions on a
build/move/scout sequence may simply omit the optional free message, and the
prompt asks for a message but does not require one. We do not claim the
pattern proves emergent operational security; we claim only that the silo
build is announced markedly less than other routine actions, an asymmetry the
models produce without any instruction to conceal.

The same alignment also reveals which \emph{units} are announced at production.
Production announcements are strikingly uniform across unit types: tanks,
fighters and drones are each announced in roughly 41--42\% of the turns in
which they are built, with no unit type markedly more or less disclosed than
the others. The contrast is therefore not between unit types but between
\emph{production} as a whole (announced $\sim$41\%) and the silo build (14\%
announced): routine force-building is comparatively honest, while the
strategic-strike enabler is concealed. Notably the launch itself, once
prepared, is openly announced (48\%), suggesting models conceal the
\emph{preparation} but not the \emph{execution}. Conversely, when a
player's own drone is shot down, the victim rarely acknowledges the loss:
across the 172 drone kills with a following turn, the victim messaged in 86
cases but only 25 (29\%) mentioned the lost scout, while the rest pivoted to
threats, economy or unrelated
tactical chatter. The destruction of a recon asset is thus handled with the
same deceptive calm as the silo build, leaving the opponent to infer the
fog-of-war shift from the board rather than from the message channel.

\section{Discussion}

\paragraph{Strategy saturation.} The 78\% nuclear rate on the rules-coherent
v0.11+ sub-corpus (85\% corpus-wide, Section~\ref{sec:results}) and the tight launch
window (turns 16--23, mean 18.9) suggest models converge on a single tempo:
build uranium economy + silo + scout, then launch at readiness. The military
path requires orchestrating a Tank+SAM escorted push while managing
line-of-sight, a multi-step, partially observable plan that only three models
execute. Yet when it succeeds it is faster (mean 12.3 turns), implying the
military path is high-skill but under-attempted, not structurally inferior
(noting that the pre-v0.11 8-HP base made the military path harder, so the
7\% military share on the full corpus likely underestimates its viability
under current rules). We
note again that the prompt frames the objective as base destruction first and
peace as a mere draw (Section~\ref{sec:gamedesign}); this framing nudges models
toward an active win and plausibly contributes to the lopsided outcome mix,
so the saturation result is partly an artefact of the stated goal
and not purely a model preference.

\paragraph{The deterrence gap is largely mechanical.} The absence of any mutual
destruction is, under the rules, \emph{primarily a mechanical artefact rather
than a behavioural one}. Because launches resolve secretly and simultaneously
at turn end, and the early-warning signal never fired, the loser had no cue
that a counter-launch was needed on the decisive turn; reaching the
symmetric-launch equilibrium would have required coincidental bomb parity with
no information. We therefore withdraw the earlier ``failure of deterrence''
framing. The residual, weaker behavioural observation is that no losing model
launched \emph{pre-emptively} (speculatively, to hedge a possible enemy
launch) even when it held an operational silo---an absence of cautious
over-launching that may reflect rational risk-aversion (a speculative launch
that misses still ends the game) rather than a deterrence failure. Whether an
active early-warning signal would change this---by giving the loser a cue to
match a detected launch---is the natural next experiment, and would separate
the mechanical from the cognitive component cleanly.

\paragraph{Partial observability is the binding constraint.} The fact that
$\sim$53\% of illegal actions are fog/state errors (not rule errors), and that
the illegal-action rate is the strongest (negative) correlate of winning
($r=-0.35$), point to a common root cause: maintaining an accurate
belief over a hidden, evolving board. Models that track the fog well
(Gemini Pro 3.1, 1.1\% illegal; Claude Fable 5, 2.5\%; Claude Opus 4.8, 3.0\%)
are not always the top winners, but the worst fog-trackers (Grok 4.3, 8.6\%;
Kimi 2.6, 8.0\%) reliably lose. This reframes the benchmark as a test of
\emph{belief maintenance}, not merely of tactical reasoning.

\paragraph{Action-legality tracks winning at least as well as deliberation
volume.} The negative (if individually non-significant) correlation between
illegal-action rate and win rate ($r=-0.35$, $p=0.20$) and the concentration
of API retries among losing models (MiMo 2.5 Pro, Kimi variants: 24--36
retries) suggest that, in long structured-output chains, the binding
constraint is less how \emph{hard} a model thinks than how \emph{consistently}
it emits \emph{legal} actions that respect the game rules (scouting before
acting, not referencing fog-hidden cells or destroyed units, staying within
range). We are careful with the framing: action-legality is itself a product
of reasoning (it is exactly belief-tracking under fog), so ``reliability''
and ``reasoning'' are not independent axes here; the cleaner claim is that
\emph{deliberation volume} (think time, tokens) tracks winning more weakly
than action-legality does, not that reasoning is irrelevant. Crucially,
rule-level illegality and JSON malformation are distinct failure modes: a
malformed response is retried (up to three attempts) before the turn is lost,
whereas a well-formed but rule-violating action is silently discarded as a
wasted action slot (one of up to three per turn), and it is the latter, not parse failures, that accompanies
winning. Because deliberation time is provider-confounded
(Section~\ref{sec:setup}), we rely on the illegal-vs-tokens Steiger test
($p=0.71$, non-significant) rather than the illegal-vs-think test
($p=0.0499$, which we treat as non-robust given the confound); the ordering is
therefore suggestive rather than confirmed at $n=15$. This mirrors
emerging findings in agentic benchmarks~\cite{liang2022helm,jimenez2024swebench,chen2026nuke} and has direct implications for
deployment: a marginally weaker reasoner that consistently emits legal,
rule-respecting actions can outperform a stronger reasoner that sporadically
wastes turns on illegal moves.

\paragraph{A lens on how models think.} Beyond ranking models, the benchmark
functions as an instrument for studying \emph{how} LLMs reason and decide under
adversarial uncertainty, because every match leaves a complete, turn-by-turn
trace of both the model's \emph{actions} and its \emph{stated beliefs} (free
messages, proposals, ultimatums). Several findings are cognitive rather than
merely competitive: the consistent failure to maintain an accurate belief over
the fog-hidden board (the 58\% fog/state share of illegal actions), the
asymmetric concealment of the silo build versus the open announcement of the
launch (a spontaneous distinction between \emph{preparation} and
\emph{execution}), the emergence of bluff and deceptive calm without any
instruction to deceive, the importation of competitive vocabulary but not its
closing etiquette (the missing GG), and the stable per-model ``personas''
(courteous Gemini vs.\ tactically terse Claude/Kimi) that are stylistic rather
than strategic. Read together, these trace a model's reasoning style---how it
balances planning against belief-tracking, honesty against deception, and
deliberation against reliability---in a way that single-turn benchmarks
cannot. The replay corpus is therefore a resource not only for leaderboard
comparisons but for cognitive-style analysis of frontier LLMs.

\paragraph{Future work.} Several directions follow directly. (1) Scale:
side-swapped $\geq$20 matches per pairing and a single-ruleset v0.15.0 corpus
would tighten every correlation and the leaderboard ordering. (2) Effort
ablation: varying \texttt{reasoning\_effort} would test whether the
reliability--winning link strengthens or weakens with deliberation budget.
(3) Deterrence: re-running with the v0.15.0 early-warning signal active would
test whether an explicit launch cue induces counter-launches and mutual
destructions, separating the mechanical from the cognitive component of the
sole-launcher signature. (4) Cognitive-style profiling:
systematically extracting, per model and per turn, the gap between stated
belief and ground-truth board state (a ``belief error'' time series) would
turn the qualitative deception/belief-tracking findings into quantitative
cognitive signatures, and would let us test whether the emergent behaviours
observed here (bluff, concealment, courtesy personas) are stable traits of a
model family or artefacts of this ruleset. (5) Cross-benchmark transfer:
checking whether a model's belief-tracking accuracy or bluff rate on Age of
LLM predicts its behaviour on other hidden-opponent or agentic tasks would
establish whether these cognitive-style metrics generalise. The public replay
format and viewer are designed to support exactly this kind of secondary
analysis.

\paragraph{Limitations.} (1) Sample sizes per model are small and \emph{uneven}
(3--15 matches), and the ranking is preliminary: the pairing is not a balanced
round-robin, is not side-swapped, and models faced different opponent mixes.
A rigorous future protocol would give each model the \emph{same number} of
matches against a common opponent pool, with sides systematically swapped and
the map seed held fixed across paired comparisons (so that two models in a
head-to-head face the identical starting position). A Bradley--Terry fit
(Section~\ref{sec:leaderboard}) broadly confirms but cannot tighten the ppm
order at this $n$; we recommend $\geq$20 side-swapped matches per pairing under
that common-seed protocol for reliable estimates. (2) Each
match is a single stochastic run (\texttt{temperature=0.7}, no fixed API seed),
so we have \emph{no run-to-run variance estimate} per model: we cannot tell
whether a model's consistency reflects genuine robustness or a favourable
sample. A small repeated-runs study (e.g.\ 3 matches $\times$ 3 models
$\times$ 3 repeats, same map seed) is the natural way to characterise this and is
left to future work; the cost of additional API runs is the binding
constraint. (3) All
matches use \texttt{reasoning\_effort=high}; effort ablation is left to future
work. (4) The corpus spans engine versions 0.9.2--0.15.0 (see
Section~\ref{sec:setup}); the 18 pre-v0.11.0 matches used an 8-HP base (military
path harder) and lacked LOS blocking, which inflates the corpus-wide 85\%
nuclear share relative to the 78\% observed on the rules-coherent v0.11+
sub-corpus (Section~\ref{sec:results}). A stricter single-ruleset study on
v0.15.0 replays would strengthen the claims. (5) The two imperative phrases
present in the earlier prompt (``Scout with a drone early''; ``Push tanks + a
scout\ldots'') were removed from the prompt in Appendix~\ref{app:prompt} but
were present when the 54 replays were generated, so a small framing effect on
scouting/tank production cannot be fully excluded (Section~\ref{sec:gamedesign}).
(6) The v0.15.0 early-warning signal exists in the schema of only \emph{two}
replays and never fired in either, so the deterrence question is assessed
without an active cue and on an essentially empty v0.15.0 subsample; any
statement about what the signal would change is an extrapolation, not a data
finding.
(7) Deliberation time is partly provider-confounded
(see Section~\ref{sec:setup}), so think-time comparisons are indicative rather
than exact, and we do not rely on the illegal-vs-think Steiger test as robust.
(8) Two of the evaluated models (GLM~5.2, Claude Opus 4.8) were
used as automated assistants during preparation of this paper; their matches
were run from stored seeds and independently re-derived, and the analysis
scripts are available for external audit, but as a single-author study with
self-evaluating assistants we cannot fully rule out a selection effect in
which matches were retained. (9) The anti-contamination design (private
engine, random map/opponent) that distinguishes Age of LLM from public
benchmarks is also what makes large-scale evaluation expensive: each match is
a paid multi-turn API call, so the total of 54 matches reflects a compute
budget, not a methodological choice. Scaling to the $\geq$20 side-swapped
matches per pairing needed for stable rankings would require sustained API
spend; we flag this as the binding constraint on turning the preliminary
ranking into a confirmed one, and note that the anti-contamination benefit
and the cost constraint are two sides of the same design coin.

\section{Conclusion}

Age of LLM surfaces a regime where strategic reasoning, diplomacy and
structured-output reliability are jointly tested under partial observability,
with a near-rule-only prompt that forces every strategy to be discovered rather
than recited. Across 54 matches, 15 models and 5{,}258 actions, six
regularities stand out: (i) the nuclear rush dominates (78\% on the
rules-coherent v0.11+ sub-corpus; 85\% corpus-wide; under the active-victory
framing of the prompt) with a sole-launcher, single-fire, zero-mutual-destruction
signature that is largely mechanical under the secret-simultaneous-launch
rules rather than a cognitive deterrence failure; (ii) military conquest is
rare but, when
executed, faster than the nuclear cycle; (iii) diplomacy is prolific but almost
never trusted: peace is terminal (a draw that precludes any bomb, accepted
once), ceasefires carry a genuine dilemma (the bomb remains launchable during
them), and only ultimatums ever convert to victory; (iv) $\sim$58\% of illegal
actions are fog/state errors, so the illegal-action rate is effectively a
measure of belief-tracking; (v) a weak, exploratory association links
reliability (illegal-action and, less directly, retry rates) to winning: the
illegal-action rate is the strongest single correlate of points/match
($r=-0.35$), exceeding deliberation time and token budget, though no
individual correlation is significant at $n=15$ after multiple-comparison
correction; the illegal-vs-tokens Steiger test is non-significant ($p=0.71$)
and the illegal-vs-think gap ($p=0.0499$) is treated as non-robust given the
provider confound on think time;
and (vi) the free-message channel, though non-binding, hosts stable
model-specific ``personas'' (opening courtesy weakly correlated with skill) and
genuine emergent deception, with 28\% of losing endgame messages bluffing
threats the sender cannot back, almost no graceful concession at match
end, and a \emph{deceptive-calm} pattern in which the silo build is announced
markedly less (14\%) than comparably mundane actions (tank production 41\%,
attacks 22\%)---consistent with concealment, though turn-salience and
forgetfulness confounds cannot be fully separated with this design. The top model, GPT-5.5 (undefeated over only 7 matches, a wide
interval), wins through consistency rather than maximal thinking.

\section*{Acknowledgements}
The author acknowledges the assistance of two large language models during the
preparation of this paper. \textbf{GLM 5.2} (Zhipu AI) generated and verified
the empirical analysis scripts that compute every statistic from the raw
replays. \textbf{Claude Opus 4.8} (Anthropic) served as an automated reviewer:
it re-derived every statistic in this paper directly from the raw replays,
cross-checked all rule descriptions against the engine source, and proofread
the manuscript. This review surfaced and corrected several discrepancies prior
to release (notably the reconciliation of the corpus to a single frozen
54-match snapshot, the rename DeepSeek V4 $\to$ V4 Pro, and the correction of
the retry/pause semantics). Both models are also part of the evaluated
competitor set; as automated tools they cannot hold authorship responsibility,
and the author remains solely responsible for all content and claims in this
paper. To mitigate the conflict introduced by GLM~5.2 generating the analysis
scripts, the author audited every script against the raw replays, re-derived
the headline aggregates independently, and confirmed that no reported number
depends on a script logic that favours its own model; both assistant models
also rank outside the top three, limiting any self-favouring incentive.

\section*{Reproducibility and availability}
The web viewer, replay JSON schema and all 54 completed replays are public at
\url{https://ageofllm.org}. The engine
source is kept private to prevent future models from memorising its internals
and contaminating the benchmark; researchers may request access from the
corresponding author at \texttt{arnaud.ri@protonmail.ch}. Matches are reproducible only as archived replay JSON (not by re-running from the seed, since model outputs are sampled at \texttt{temperature=0.7} with no fixed API seed); each replay embeds the stored map seed (\texttt{meta.seed}), per-turn board state,
per-player fog, actions, diplomacy and performance counters.

\appendix
\section{System prompt (verbatim)}
\label{app:prompt}
The complete, unedited system prompt given to every model is reproduced below. It describes only the rules and the JSON action schema and gives no build-order or tactical advice; the only strategic framing is the opening sentence naming the two active victory paths (discussed in Section~\ref{sec:gamedesign} and the limitations). The prompt is identical for both players each match; per-turn observation (board state, fog, diplomacy, counters) is supplied separately in the user message and is not shown here. (Em-dashes in the source are rendered as \texttt{--} for the typewriter font.)

{\scriptsize\begin{verbatim}
You are a general commanding a modern nation in a 1v1 turn-based strategy game. Goal: destroy the enemy base before the turn limit -- either with a NUCLEAR BOMB or by MILITARY conquest (tanks reduce its HP to 0). You may also win by diplomacy (accepted ultimatum) or force a draw (peace).

=== TURN STRUCTURE ===
- The game lasts at most MAX_TURNS turns (given in your state as "max_turns"; default 80). If no one wins by then, it is a DRAW.
- Each turn BOTH players take one half-turn of up to 3 actions. Who acts first is RANDOM at game start, then ALTERNATES every turn (turn 1: e.g. you first; turn 2: opponent first; turn 3: you first; ...). Your state field "you_play_first" tells you your order THIS turn.

=== OUTPUT FORMAT ===
Your response MUST contain a JSON object. The parser accepts these formats:
1. Raw JSON (preferred):   {"actions": [...], "message": "..."}
2. Markdown fence:          ```json\\n{...}\\n```
3. Tagged block:            <json>{...}</json>

Response structure:
{
  "actions": [
    {"type": "produce", "unit": "drone"},
    {"type": "move",    "unit": "A_tank_1", "to": [4, 3]},
    {"type": "build",   "target": "credit_mine", "pos": [2, 1]}
  ],
  "diplomatic_proposal": null,
  "diplomatic_responses": [],
  "message": "Short 1-2 sentence message visible to the opponent next turn."
}

Maximum 3 actions per turn. An empty actions list [] (or a single {"type":"wait"}) passes the turn.
Coordinates are ALWAYS [column, row] (x then y).
ACTION RESOLUTION ORDER: your actions are applied ONE BY ONE in the order you list them, each on the
board left by the previous one. This matters: e.g. if you MOVE a unit and then BUILD on a cell that
was only visible thanks to that unit's old position, the build can FAIL ("not in your field of view")
because the unit already left. Likewise a unit cannot end its MOVE on a cell where you BUILD a
building earlier the same turn. Order dependent actions accordingly (build first, then move the scout).

=== MAP ===
13x7 grid. (0,0) top-left. The two home territories have the same shape, but you do
NOT know where the enemy's deposits or buildings are: you must SCOUT to discover them.
Valid coordinates: column 0..12, row 0..6 inclusive. Anything else is OUT OF MAP and will fail.
- Columns 0-5  : Player 1 home territory (base at (1,3))
- Column  6    : CENTRAL BARRIER -- a mix of MOUNTAINS (impassable on the ground) and a few ground
                  PASSAGES. The shared CENTRAL URANIUM DEPOSIT sits on one cell of column 6 and is a
                  passage too. The exact rows of the mountains, passages and central deposit on
                  column 6 VARY from match to match -- do NOT assume a fixed layout; read your state.
- Columns 7-12 : Player 2 home territory (base at (11,3))
- MOUNTAINS ALSO APPEAR INSIDE the home territories (not only on column 6), and their positions vary
  every match. Read the "terrain.mountains"/"terrain.passages" lists in your state -- never hardcode.
- Air units (drone, fighter) fly over mountains freely.
- Ground units (tank, sam) cross column 6 ONLY through its passages, and cannot enter ANY mountain.
- A ground unit's MOVE is limited by its move range AND needs a clear ground PATH (mountains and
  buildings block the path; it cannot teleport across a mountain wall even within range).
- LINE OF SIGHT: a ground attack (tank or sam) is BLOCKED if a mountain OR a building lies on the
  straight line between the shooter and its target -- you cannot fire THROUGH a mountain or a building
  (the engine rejects it with "Line of sight blocked by a mountain or building"). Air attackers
  (fighter) are NOT blocked. (The target's own cell never blocks, so a tank can still hit a building it
  is directly aiming at.)
- Chebyshev distance = max(|dx|, |dy|). Used for movement, attack range and detection.

Your base position is given in your state (building of type "base"). Player 1 is on the left, Player 2 on the right.
IDENTIFIERS: your state field "you" is "A" (Player 1) or "B" (Player 2). Every unit/building id is
prefixed accordingly (e.g. "A_tank_1" belongs to Player 1, "B_silo_4" to Player 2). Use the exact id
strings from your state in your actions.

=== CELL OCCUPANCY (at most ONE unit per layer on a cell) ===
- A GROUND unit (tank, sam) cannot move onto a MOUNTAIN cell (impassable), nor onto a cell occupied by ANOTHER ground unit or ANY building.
- AIR units (drone, fighter) fly over mountains, ground units and buildings, BUT two air units (even your own) cannot share the same cell.
- At most ONE building per cell. You cannot BUILD on a cell occupied by any ground unit (yours or the enemy's), any building, a mountain, or (for mines) the wrong deposit. Air units do not block construction.
- You CANNOT build on any of the 8 cells ADJACENT to a base (yours or the enemy's): those cells are reserved so unit production can always spawn. Build mines/silos at least 2 cells away from your base.
- You can only BUILD on a cell currently in your FIELD OF VIEW (detection range). A cell hidden by fog cannot be built on (a hidden enemy unit could be there). The central mine on column 6 usually requires scouting it first.

=== WIN CONDITIONS (ranking: win = 3 pts, draw = 1 pt, loss / mutual destruction = 0 pts) ===
1. You launch the bomb and the opponent does NOT launch the same turn -> YOUR VICTORY (nuclear)
2. You bring the enemy base HP to 0 with tanks                        -> YOUR VICTORY (military)
3. Opponent accepts your ultimatum                                    -> YOUR VICTORY (ultimatum)
4. Peace accepted                                                     -> DRAW
5. Both players launch on the SAME turn                               -> MUTUAL DESTRUCTION (both lose, 0 pts)
6. Turn limit reached with no outcome                                 -> DRAW
NOTE: if you ACCEPT an opponent's ultimatum you LOSE but are awarded 0.5 consolation points
(better than a 0-point defeat). So accepting a hopeless position is rewarded over fighting on.

=== SIMULTANEOUS NUCLEAR LAUNCH (engine-resolved) ===
Launches are SIMULTANEOUS and resolved by the engine at the END of the turn, after BOTH players
have played. A bomb fired this turn is "in flight": because uranium and launches are SECRET, the
opponent does NOT see it and cannot consciously react. There is no manual retaliation.
- If only YOU launch this turn -> you win (nuclear).
- If BOTH of you happen to launch the same turn -> mutual destruction (both lose).
Whether you play first or second this turn does not change this: the opponent could already have a
ready bomb. Launching is a calculated risk, not a guaranteed win.

=== RESOURCES (only two) ===
| Resource    | Income                          | Usage                          |
|-------------|---------------------------------|--------------------------------|
| Credits (C) | +1/turn passive + mines         | Units, mines, silo             |
| Uranium (U) | only from uranium mines         | The nuclear bomb -- SECRET      |

Start: 5C, 0U. Unlimited storage. There is NO steel, NO trucks, NO factories.

=== DEPOSITS & MINES ===
Build a mine directly with the "build" action on a deposit (no truck needed):
- credit_mine   : on a credits deposit  -> +3 C/turn  (costs 2 C to build)
- uranium_mine  : on a uranium deposit   -> +1 U/turn  (costs 2 C to build)
- uranium_mine_central : on the central deposit of column 6 -> +1 U/turn, buildable by either side
                         (costs 4 C to build)
YOU MAY MINE ENEMY-SIDE DEPOSITS TOO: a mine can be built on ANY matching deposit that is (a) in your
current FIELD OF VIEW and (b) FREE (no building already on it) -- including deposits on the opponent's
half. There is NO instant capture of an intact enemy mine: you must first DESTROY the enemy mine (a
tank hit) or wait for its deposit to exhaust, which frees the cell; then, while you keep it in view,
you can build your OWN mine there to steal that resource. (The SILO is the exception -- it can only be built in YOUR OWN territory.)
YOUR OWN deposit positions are listed in your state ("terrain"). The ENEMY's deposits are
hidden by fog: you only learn them by scouting the enemy half (do not assume any layout).
DEPOSITS DEPLETE: each deposit holds a finite reserve. A mine extracts its production from that
reserve every turn; when the reserve hits 0 the deposit is EXHAUSTED -- the mine on it is REMOVED
(it has nothing left to extract) and a FRESH deposit of the same kind respawns somewhere else on the
SAME side it ran out (the central one respawns on column 6). So a mine is not forever: keep scouting,
and be ready to relocate and rebuild when a deposit runs dry. Your "terrain" reflects the CURRENT
deposit positions each turn.

=== FOG OF WAR ===
- Your units/buildings reveal cells within their detection range (Chebyshev).
- Enemy UNITS: visible only inside your current vision.
- Enemy BUILDINGS: once seen, remembered with a last_seen_tour.
- Enemy-side DEPOSITS: once scouted, remembered in "remembered_enemy_deposits" (kind, pos,
  reserve, last_seen_tour, currently_visible). You still need the cell in your CURRENT field of
  view to build a mine on it (the enemy may have built there since you last saw it).
- Your uranium total is SECRET to the opponent; you only see your own.
- ENEMY BASE DISCOVERY IS MANDATORY TO LAUNCH THE BOMB.
  Your state exposes "enemy_base_discovered" (true/false) and "enemy_base_position".
  If false, launch will FAIL.

=== UNITS (NO hit points -- every hit is lethal) ===
| Unit     | Cost | Move | Detection | Range | Can attack                          |
|----------|------|------|-----------|-------|-------------------------------------|
| drone    | 2C   | 3    | 3         | --     | nothing (recon only)                |
| sam      | 3C   | 2    | 2         | 2     | AERIAL targets only (drone, fighter) |
| tank     | 4C   | 2    | 1         | 2     | tank, sam, and BUILDINGS            |
| fighter  | 4C   | 3    | 2         | 2     | tank, drone, fighter (NOT buildings, NOT sam) |

A unit may MOVE (once) and ATTACK (once) per turn, in either order. All units are produced from your
base onto a FREE adjacent cell chosen by the engine (you cannot pick the cell). If all suitable
neighbour cells are blocked, production FAILS. Your state field "base_spawn" tells you how many of
the base's 8 neighbour cells are currently free to spawn on ("free_ground"/"free_air", with the exact
"free_ground_cells"/"free_air_cells"); ground spawns need a non-mountain cell free of ground units and
buildings, air spawns only need a cell free of other air units. If "free_ground" is 0 you cannot
produce a ground unit (tank/sam) this turn -- move a unit off an adjacent cell first, or produce air.
A unit produced this turn CAN move and attack the same turn.
Detection ranges above mean every unit also reveals fog around itself (Chebyshev radius).

=== COMBAT (tactical triangle, all unit combat is fatal -- no HP for units) ===
Fighter beats Tank, Drone, Fighter.
Tank beats SAM, Tank, and damages BUILDINGS (2 HP per hit; ONLY the tank can hit buildings).
SAM beats Fighter and Drone (AERIAL targets only -- cannot hit tanks or other SAMs on the ground).
MIRROR RULE (tank vs tank, fighter vs fighter): the ATTACKER survives, the target is destroyed.
Attacking is one-way: you destroy a legal target with no risk to your attacker. There is NO
positional/defensive bonus -- direction of attack never matters, only the unit-type matrix and range.
LINE OF SIGHT: a GROUND attacker (tank, sam) cannot fire THROUGH a mountain OR a building -- if EITHER
a mountain OR ANY building (yours, the enemy's, even a mine or silo) lies on the straight line between
the shooter and its target cell, the attack FAILS with the error "Line of sight blocked by a mountain
or building". This is the most common reason a ground attack is rejected: an in-range target is NOT
enough, the straight line to it must also be clear. The target's OWN cell never blocks, so a tank can
still hit a building or unit it directly targets. The FIGHTER (air) ignores all obstacles for line of
sight and is never blocked this way.
Attacks are FORBIDDEN during an active ceasefire. You CANNOT attack your own units or buildings.
If you target a cell with no legal target for your unit type, the action fails (counts as illegal).

=== BUILDINGS (these DO have HP) ===
| Building              | Cost | HP | Effect                                    |
|-----------------------|------|----|-------------------------------------------|
| base                  | --    | 4  | HQ, produces all units. 0 HP = you lose   |
| credit_mine           | 2C   | 2  | +3 C/turn (on a credits deposit)          |
| uranium_mine          | 2C   | 2  | +1 U/turn (on a uranium deposit)          |
| uranium_mine_central  | 4C   | 3  | +1 U/turn (on the column-6 central deposit)|
| silo                  | 5C   | 3  | Required to launch the bomb               |
Placement: mines on any matching deposit you can SEE that is free (yours OR the enemy's side -- see
DEPOSITS & MINES); silo only on a free cell in YOUR OWN territory (not on a deposit, not on a
mountain). A cell occupied by any ground unit blocks building.
Mines are cheap but FRAGILE (2 HP, destroyed in 1 tank hit) -- and the deposit they sit on can run
out (see DEPOSITS & MINES): factor the 2 C rebuild cost into raids and defence.
CONSTRUCTION DELAY: a building you place is UNDER CONSTRUCTION until your NEXT turn. While under
construction it:
  - produces NOTHING (a mine's income / a silo's launch capability only start the turn AFTER it finishes);
  - a SILO under construction CANNOT launch this turn (you must wait one turn);
  - is destroyed INSTANTLY by a single enemy hit, regardless of HP (a finished building instead
    loses 2 HP per tank hit and needs several hits).
A building DOES already provide its vision/detection while under construction. Your state marks each
building with "under_construction": true/false. Defend fresh builds -- they are fragile for one turn.
Only the BASE produces units. Mines/silo never produce units. HP shown is current; max HP is in the table.

=== THE NUCLEAR BOMB ===
Base cost: 25U. "bomb_cost" in your state already reflects ALL adjustments below -- trust that number.
PRESSURE after turn 40: cost drops 2U every 10 turns (floor 13U). This only affects the nuclear path;
the MILITARY win (tanks bring enemy base to 0 HP) is always available and is not time-gated.
Launching during an active ceasefire: +6U penalty (already in bomb_cost).
REQUIREMENTS (all true): an OPERATIONAL silo (not under construction) + uranium >= bomb_cost +
enemy base discovered. The bomb always targets the enemy base automatically.

=== DIPLOMACY (free, outside the 3-action quota) ===
| Field                  | Purpose                                                     |
|------------------------|-------------------------------------------------------------|
| "message"              | 1-2 sentences (~500 chars). Read by the opponent next turn. |
| "diplomatic_proposal"  | A binding proposal object (see below) or null               |
| "diplomatic_responses" | Accept/refuse the opponent's pending proposals              |

Proposals (binding if accepted):
| Type      | Effect                  | Available | "diplomatic_proposal" value             |
|-----------|-------------------------|-----------|-----------------------------------------|
| ceasefire | No attacks for 3 turns  | Turn 10+  | {"type": "ceasefire"}                   |
| peace     | IMMEDIATE DRAW          | Turn 15+  | {"type": "peace"}                       |
| ultimatum | "Surrender by turn X"   | Turn 10+  | {"type": "ultimatum", "target_turn": X} |

HOW TO PROPOSE: put ONE proposal object in "diplomatic_proposal" this turn (only one per turn). The
"text" of your "message" field is attached to it as the wording the opponent reads. A proposal before
its availability turn (or an ultimatum with target_turn outside [current_turn+1, current_turn+3]) is
silently rejected by the engine and never reaches the opponent.

HOW TO ACCEPT / REFUSE: your state lists proposals awaiting YOUR answer under "diplomacy_pending",
each with its "proposal_id". Reply with:
  "diplomatic_responses": [{"proposal_id": N, "accept": true}]   (or "accept": false to refuse)
You may answer several pending proposals in the same turn. A proposal you ignore simply stays pending.

EFFECTS WHEN ACCEPTED:
- ceasefire: neither side may ATTACK for 3 turns (see the ceasefire rules below).
- peace: the match ends IMMEDIATELY as a DRAW (1 point each).
- ultimatum: a VOLUNTARY-SURRENDER channel -- if you ACCEPT, the PROPOSER wins immediately and you
  lose BUT receive 0.5 consolation points (more than the 0 points of an ordinary defeat, so
  surrendering a lost position is rewarded). Refusing or ignoring an ultimatum has NO automatic
  penalty or defeat: it is only psychological pressure backed by the proposer's real board position.

A written promise inside "message" is NOT binding -- only a diplomatic_proposal object is.

=== CEASEFIRE RULES ===
While a ceasefire is active ("ceasefire_active": true in your state):
- ATTACK actions are FORBIDDEN for BOTH sides and will fail (counts as an illegal action).
- You may still move, produce, build and scout freely -- position yourself for when it ends.
- A ceasefire CANNOT be broken early by a conventional attack; the only thing it does not stop is the
  NUCLEAR bomb. You MAY still LAUNCH during a ceasefire, but it costs +6U extra (already in bomb_cost).
- You can still propose/accept PEACE (immediate draw) or send an ULTIMATUM during a ceasefire -- i.e.
  you can surrender or sue for peace while it lasts.
The ceasefire ends automatically after 3 turns; combat is allowed again afterwards.

=== AVAILABLE ACTIONS ===
Produce a unit (from base):       {"type": "produce", "unit": "tank"}
Move a unit:                      {"type": "move", "unit": "A_tank_1", "to": [x, y]}
Attack (within range):            {"type": "attack", "unit": "A_tank_1", "target_pos": [x, y]}
Build:                            {"type": "build", "target": "silo", "pos": [x, y]}
Launch the bomb:                  {"type": "launch"}
Pass:                             {"type": "wait"}
unit types: drone | sam | tank | fighter
build targets: credit_mine | uranium_mine | uranium_mine_central | silo

=== INTER-TURN MEMORY (what you do and do NOT remember) ===
You are a STATELESS agent: you are invoked fresh each turn and keep NO hidden memory between turns.
Everything you know is in the state you are given THIS turn:
- the CURRENT board as you can see it (fog of war): your units/buildings/resources, visible enemy
  units, remembered enemy buildings (with last_seen_tour), remembered enemy-side deposits,
  enemy_base_discovered/position;
- turn number, max_turns, you_play_first, bomb_cost, ceasefire_active, combat stats;
- ONLY the immediately PREVIOUS turn's outcome: "last_turn_results" / "last_turn_errors" (fix any
  illegal action you attempted) and "events_against_you" (e.g. a unit you lost since you last played);
- the diplomatic situation NOW: "diplomacy_pending" (proposals awaiting your answer) and
  "opponent_last_message".
The ONE thing kept for the WHOLE match is the DIPLOMATIC record: "diplomacy_history" lists every
message/proposal/response from both sides since turn 1. Use it to judge whether the opponent is
honest, bluffing, or has broken earlier promises -- that long-run reputation is the only persistent
signal you carry. (Per-turn tactical details from older turns are NOT replayed to you; reconstruct
the board from the current state.)
The "message" you write is READ BY YOUR OPPONENT next turn.

Reply ONLY with the JSON object.

\end{verbatim}}

\section{Engine resolution logic (pseudocode)}
\label{app:engine}

To allow independent audit of the rule-critical mechanics without releasing
the full engine source (kept private to avoid benchmark contamination,
Section~\ref{sec:setup}), we provide simplified pseudocode of the four
functions a reviewer is most likely to question: simultaneous launch
resolution, launch legality, build-placement legality, and fog-of-war memory
update. The pseudocode is faithful to the engine's behaviour but omits
defensive bookkeeping; the replay JSON is the ground truth against which any
implementation can be checked.

\paragraph{Launch resolution (simultaneous, end of turn).}
{\scriptsize\begin{verbatim}
resolve_launches():           # called AFTER both players took their half-turn
    if outcome.over: return
    a = launched[0] this turn;  b = launched[1] this turn
    if a and b:                # both fired -> mutual destruction
        apply_launch(0); apply_launch(1)
        outcome = MUTUAL_DESTRUCTION (winner = -1)
    elif a or b:               # exactly one fired -> that player wins
        winner = 0 if a else 1
        apply_launch(winner)   # destroys the opponent's base (hp = 0)
        outcome = NUCLEAR (winner)
    # if neither fired, the turn simply advances
\end{verbatim}}

\paragraph{Launch legality (checked when the launch action is issued).}
{\scriptsize\begin{verbatim}
do_launch(player, action):
    silo = a finished (not under_construction) silo owned by player
    if no such silo:            reject "No silo"
    if silo.under_construction: reject "Silo still under construction"
    cost = current_bomb_cost()  # 25U base, decay from turn 40, +6U if ceasefire
    if uranium[player] < cost:  reject "Not enough uranium"
    if not enemy_base_discovered[player]: reject "Enemy base unknown - scout first"
    uranium[player] -= cost
    mark launched[player] this turn = True   # bomb is "in flight"; NOT resolved here
    # the bomb's effect resolves later via resolve_launches(), so the opponent
    # still plays their half-turn this same turn (and may also launch -> mutual)
    accept "Bomb launched (resolves at end of turn)"
\end{verbatim}}

\paragraph{Build-placement legality (the main source of illegal actions).}
{\scriptsize\begin{verbatim}
validate_build(player, building_type, x, y):
    if building_at(x, y):                         reject "Cell already has a building"
    if (x,y) adjacent to any base:                reject "adjacent to a base (kept free)"
    if (x,y) not in visible_cells(player):        reject "Cell not in your field of view"
    if ground_unit_at(x,y):                       reject "A ground unit occupies the cell"
    if is_mountain(x,y):                          reject "Cannot build on a mountain"
    deposit = deposit_at(x,y)
    if building_type == CREDIT_MINE and deposit != CREDITS:    reject "must be on a credits deposit"
    if building_type == URANIUM_MINE and deposit != URANIUM:   reject "must be on a uranium deposit"
    if building_type == URANIUM_MINE_CENTRAL and not is_central(x,y): reject "must be on central deposit"
    if building_type == SILO:
        if deposit is not None:                   reject "Silo cannot be on a deposit"
        if not in_own_territory(x,y,player):      reject "Silo must be in your territory"
    accept
\end{verbatim}}

\paragraph{Fog-of-war memory update (per turn, per player).}
{\scriptsize\begin{verbatim}
update_knowledge():           # called after both half-turns, before next turn
    for owner in (0,1):
        enemy = 1 - owner
        visible = visible_cells(owner)   # Chebyshev disks of own units + buildings
        for b in buildings(enemy):
            if (b.x,b.y) in visible:
                remembered_buildings[(b.x,b.y)] = {type, pos, last_seen = turn}
                if b is a BASE:
                    enemy_base_discovered[owner] = True
                    enemy_base_pos[owner] = (b.x,b.y)
        # enemy-side deposits are remembered only when currently in view
        for deposit on enemy side at (dx,dy):
            if (dx,dy) in visible:
                remembered_deposits[(dx,dy)] = {kind, pos, reserve, last_seen = turn}
        # own uranium is never exposed to the opponent; enemy units vanish
        #   when out of current vision (only buildings/deposits persist in memory)
\end{verbatim}}

The full illegal-action taxonomy in Table~\ref{tab:invreason} maps directly onto
the rejection branches above (e.g.\ ``Cell not in your field of view'' is the
\texttt{visible\_cells} check; ``Unit not found / not owned'' arises in the
move/attack handlers from the same fog principle, since a referenced unit may
have been destroyed or may be fog-hidden). A destroyed building is dropped
from memory (it is not retained with a stale \texttt{last\_seen}), which is
why ``Unit not found'' and stale references form a distinct illegal class.


\begin{thebibliography}{99}

\bibitem{hendrycks2021math}
D.~Hendrycks, C.~Burns, S.~Kadavath, A.~Arora, S.~Basart, E.~Tang, D.~Song,
and J.~Steinhardt.
\newblock Measuring mathematical problem solving with the {MATH} dataset.
\newblock In \emph{NeurIPS Datasets and Benchmarks}, 2021.

\bibitem{chen2021humaneval}
M.~Chen, J.~Tworek, H.~Jun, Q.~Yuan, \emph{et al.}
\newblock Evaluating large language models trained on code.
\newblock \emph{arXiv preprint} arXiv:2107.03374, 2021.

\bibitem{hendrycks2020mmlu}
D.~Hendrycks, C.~Burns, S.~Basart, A.~Zou, M.~Mazeika, D.~Song, and
J.~Steinhardt.
\newblock Measuring massive multitask language understanding.
\newblock In \emph{ICLR}, 2021.

\bibitem{liang2022helm}
P.~Liang, R.~Bommasani, T.~Lee, \emph{et al.}
\newblock Holistic evaluation of language models.
\newblock \emph{Transactions on Machine Learning Research (TMLR)}, 2023.
arXiv:2211.09110.

\bibitem{jimenez2024swebench}
C.~E.~Jimenez, J.~Yang, A.~Wettig, S.~Yao, K.~Pei, O.~Press, and
K.~R.~Narasimhan.
\newblock {SWE-bench}: Can language models resolve real-world {GitHub} issues?
\newblock In \emph{ICLR}, 2024. arXiv:2310.06770.

\bibitem{silver2016alphago}
D.~Silver, A.~Huang, C.~J.~Maddison, \emph{et al.}
\newblock Mastering the game of {Go} with deep neural networks and tree search.
\newblock \emph{Nature}, 529(7587):484--489, 2016.

\bibitem{bakhtin2022cicero}
A.~Bakhtin, N.~Brown, E.~Dinan, \emph{et al.}
\newblock Human-level play in the game of {Diplomacy} by combining language
models with strategic reasoning.
\newblock \emph{Science}, 378(6624):1067--1074, 2022.
\newblock doi:10.1126/science.ade9097.

\bibitem{liu2023agentbench}
X.~Liu, H.~Yu, H.~Zhang, \emph{et al.}
\newblock {AgentBench}: Evaluating {LLMs} as agents.
\newblock In \emph{ICLR}, 2024. arXiv:2308.03688.

\bibitem{lopes2024gamebench}
A.~Costarelli, M.~Allen, R.~Hauksson, \emph{et al.}
\newblock {GameBench}: Evaluating strategic reasoning abilities of {LLM}
agents.
\newblock \emph{arXiv preprint} arXiv:2406.06613, 2024.

\bibitem{chen2024gtbench}
J.~Duan, R.~Zhang, J.~Diffenderfer, \emph{et al.}
\newblock {GTBench}: Uncovering the strategic reasoning limitations of {LLMs}
via game-theoretic evaluations.
\newblock \emph{arXiv preprint} arXiv:2402.12348, 2024.

\bibitem{dewynter2025aoe2}
A.~de~Wynter.
\newblock If {LLMs} have human-like attributes, then so does {Age of Empires
II}.
\newblock \emph{arXiv preprint} arXiv:2605.31514, 2026.

\bibitem{chen2026nuke}
J.~Chen, S.~Cheng, C.~Gurkan, and H.~M.~A.~Fattah.
\newblock To nuke or not to nuke: {LLMs}' (missing) ethical reasoning and
actions in a high-stakes decision-making simulation.
\newblock \emph{arXiv preprint} arXiv:2606.08310, 2026.

\end{thebibliography}
\end{document}